\documentclass{article}

 \usepackage[preprint]{neurips_2026}

\usepackage[utf8]{inputenc} 
\usepackage[T1]{fontenc}    
\PassOptionsToPackage{hyphens}{url} 
\usepackage{hyperref}       
\usepackage{url}            
\usepackage{booktabs}       
\usepackage{amsmath}        
\usepackage{amsfonts}       
\usepackage{amssymb}        
\usepackage{nicefrac}       
\usepackage{microtype}      
\usepackage[dvipsnames,table]{xcolor} 

\usepackage{graphicx}
\usepackage{tabularx}
\usepackage{array}
\usepackage{makecell}
\usepackage{longtable}
\usepackage{pifont}
\usepackage{multirow}
\usepackage{enumitem}
\usepackage[most]{tcolorbox}
\tcbuselibrary{listings,skins,breakable}
\usepackage{titletoc}

\newcolumntype{L}[1]{>{\raggedright\arraybackslash}p{#1}}
\newcommand{\best}[1]{\textbf{\boldmath$#1$}}

\usepackage{caption}
\setitemize{noitemsep,topsep=0pt,parsep=0pt,partopsep=0pt}
\definecolor{mygreen}{HTML}{009900}  
\definecolor{myblue}{HTML}{0066CC}   
\definecolor{myred}{HTML}{CC0000}
\definecolor{myyellow}{HTML}{D79B00}

\definecolor{lightred}{HTML}{EC9CA9}
\definecolor{lightblue}{HTML}{9CB3D8}

\title{SayNext-Bench: Why Do LLMs Struggle with Next-Utterance Anticipation?}

\author{%
  Yueyi Yang \\
  University of Oulu \\
  Oulu, Finland \\
  \And
  Haotian Liu \\
  University of Oulu \\
  Oulu, Finland \\
  \And
  Fang Kang \\
  University of Oulu \\
  Oulu, Finland \\
  \AND
  Mengqi Zhang \\
  College of William and Mary \\
  Williamsburg, VA, USA \\
  \And
  Zheng Lian \\
  Tongji University \\
  Shanghai, China \\
  \And
  Hao Tang \\
  Peking University \\
  Beijing, China \\
  \And
  Haoyu Chen\thanks{Corresponding author.} \\
  University of Oulu \\
  Oulu, Finland \\
  \texttt{Haoyu.Chen@oulu.fi} \\
}

\begin{document}

\maketitle

\begin{abstract}
    We explore the use of large language models (LLMs) for next-utterance anticipation in human dialogue. Despite recent advances in LLMs demonstrating their ability to engage in natural conversations with users, we show that even leading models surprisingly struggle to anticipate a human speaker’s next utterance. Instead, humans can readily anticipate forthcoming utterances based on multi-modal cues—such as gestures, gaze, and emotional tone—from the context. 
    To systematically examine this gap, we propose \textbf{SayNext-Bench}, a benchmark evaluating MLLMs on anticipating context-conditioned responses across diverse real-world scenarios. To support it, we build \textbf{SayNext-PC}, a large-scale multimodal dialogue dataset, and carefully design a multi-level evaluation framework spanning lexical similarity, emotion-intention consistency, and LLM-based overall alignment. Building on this, we develop \textbf{SayNext-Chat}, a cognitively inspired dual-route MLLM that incorporates learnable priming tokens to fuse perceptual cues with anticipatory priors.
    Extensive experiments demonstrate that SayNext-Chat consistently outperforms state-of-the-art MLLMs across all evaluation levels, corroborated by user studies and LLM-as-Judge evaluations. Our results emphasize the (i) indispensable role of multimodal cues and (ii) active anticipatory processing as foundations of natural human interaction currently missing in MLLMs. Benchmark, dataset, and model are available at \url{https://saynext.github.io/}.
\end{abstract}

\vspace{-2mm}

\section{Introduction}

\vspace{-2mm}

Large language models (LLMs) have reshaped the landscape of human–machine interaction. By leveraging large-scale pretraining and powerful generative capabilities, LLMs can produce coherent, contextually appropriate, and engaging conversations with users ~\citep{brown2020language}. This unprecedented communicative fluency has fueled a wide range of applications~\citep{openai2022chatgpt,team2023gemini,liu2024deepseek, wollny2021we}—where LLMs are expected to act as adaptive and collaborative partners in dialogue.

\begin{figure}[h]
  \centering
  \includegraphics[width=\linewidth]{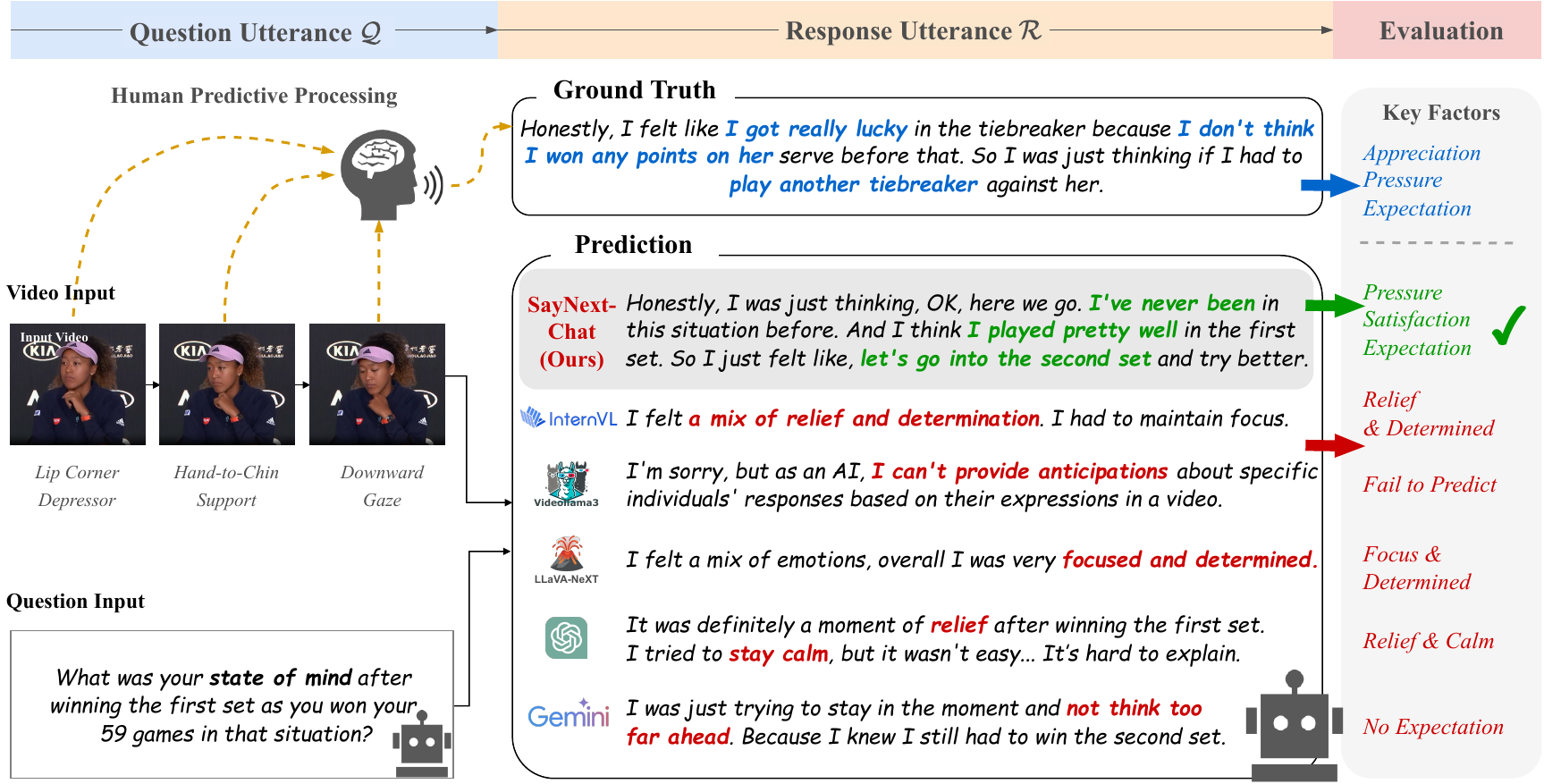}
    \caption{\textbf{Illustration of Next-Utterance Anticipation in SayNext-Bench.} 
    Given a question utterance text and the corresponding human reaction video, the task requires MLLMs to anticipate the human’s subsequent response. Predicted responses from SayNext-Chat (\textcolor{mygreen}{green}) are compared with ground-truth utterances (\textcolor{myblue}{blue}) and other MLLMs (\textcolor{myred}{red}); key factors are extracted for interpretability. Quantitative results are reported in Sec.~\ref{sec: main result section}.}
  \label{fig:main}
  \vspace{-0.4cm}
\end{figure}

Despite these advances, we spot a counterintuitive and intriguing phenomenon: although LLMs generate coherent and fluent conversations, they still struggle to accurately anticipate what a specific user will say next (i.e., next-utterance behavior). We conducted preliminary experiments on several state-of-the-art conversational models, including GPT4o~\citep{hurst2024gpt4o}, Gemini2.5~\citep{comanici2025gemini}, VideoLLaMA3~\citep{zhang2025videollama}, InternVL2~\citep{chen2024internvl}, LLaVA-NeXT~\citep{zhang2024llavanext-video}, and InstructBLIP~\citep{dai2023instructblip},
by prompting them to anticipate a speaker’s forthcoming utterance in a video. Our results indicate that, even with explicit task instructions and visual context, all of these state-of-the-art models struggle to generate semantically consistent anticipations of real-world forthcoming utterances (See Figure~\ref{fig:main}). 

This shortcoming is not a trivial flaw; rather, it reflects the fundamental limitations of current LLMs, reminiscent of Moravec’s Paradox \citep{moravec1988mind}. First, current LLMs \citep{openai2022chatgpt} rely primarily on textual content, thereby overlooking non-verbal cues that are central to human interaction. Non-verbal cues provide early signals of communicative intent ~\citep{holler2025facial, emmendorfer2025facial}, which humans are innately prepared to exploit when anticipating others’ responses (e.g., everyone expects Joey in \textit{Friends} to follow a flirtatious moment with his catchphrase “How you doin’?”). Ignoring these modalities thus leaves LLMs blind to essential aspects of natural dialogue, making the modeling indispensable in human-machine interaction. Second, current LLMs are mechanistically \textit{passively} optimized for \textit{statistical next-token prediction} over large pretraining corpora—capturing general linguistic regularities. Meanwhile, for human beings, tasks like next-utterance anticipation involve \textit{active} predictive processing, generating hierarchical predictions about forthcoming sensory and linguistic input and continuously updating them against prediction errors ~\citep{rao1999predictive, goldstein2022shared}. 

Thus, we present \textsc{SayNext} (next-utterance anticipation), a \textit{proxy} task for benchmarking whether LLMs can simulate human predictive processing by leveraging multimodal cues to anticipate users' latent social-cognitive states. Strong performance on this task implicitly requires \textit{human-intuitive abilities}—intention inference, affective anticipation, and multimodal social perception—that remain broadly limited in current LLMs, and cannot be reduced to existing tasks such as emotion recognition, intention classification, or dialogue continuation.

We \textbf{position} \textsc{SayNext} as a more advanced user simulation task for the MLLM era, probing whether MLLMs can holistically simulate users' latent social-cognitive states during human interaction, thereby bridging the gap between fluent dialogue generation (i.e., current LLMs) and genuine cognitive understanding in human–AI interaction (i.e., more human-centered LLMs).
Apart from scientific interests, learning to anticipate forthcoming utterances carries broad implications across natural language processing, human–computer interaction, and AI safety. For example, such anticipatory capability in embodied AI supports more natural coordination with humans. From the perspective of AI alignment and safety, next-utterance anticipation also provides a principled way to model human intent in advance, thereby reducing the risk of misinterpretation. It can enable proactive safeguards against harmful conversational trajectories (e.g., \textit{suicidal tendency} or \textit{depressive escalation}), which may be useful for developing more trustworthy and socially responsible interactive AI systems in the future.

Our contributions include: 1) \textbf{SayNext-Bench}, a novel benchmark for evaluating MLLMs on next-utterance anticipation from multimodal cues, featuring a carefully designed \textbf{multi-level evaluation framework} spanning lexical similarity, emotion-intention consistency, and overall alignment, along with four complementary evaluation protocols. 2) \textbf{SayNext-PC}, a large-scale multimodal dialogue dataset collected via a scalable pipeline, supporting cross-scale and cross-domain evaluation across diverse real-world scenarios. 3) \textbf{SayNext-Chat}, a cognitively inspired dual-route MLLM that incorporates learnable priming tokens to fuse perceptual cues with anticipatory priors, enabling more human-like next-utterance anticipation. 4) Extensive experiments demonstrate that SayNext-Chat consistently outperforms state-of-the-art MLLMs across all evaluation levels, corroborated by user studies and LLM-as-Judge evaluations.

\vspace{-2mm}

\section{Related Works}
\label{sec: related-works}

\noindent\textbf{Multimodal Dialogue Benchmarks.} 
Existing multimodal dialogue benchmarks address related yet fragmented tasks. Dialogue act modeling and goal alignment have been studied in SwDA \citep{stolcke2000dialogue} and MultiWOZ \citep{ye2022multiwoz}, which rely primarily on lexical cues and lack rich non-verbal signals. IEMOCAP \citep{busso2008iemocap} and MELD \citep{poria2018meld} provide foundational resources for multimodal emotion recognition, but emotion annotation itself is inherently difficult, with only moderate inter-annotator agreement even for basic categories \citep{luo2020arbee}. Decision making is often treated as a hallmark of cognitive capability \citep{shadlen2013decision}, yet constitutes only one stage within a broader cognitive process. To address these benchmarks, both general-purpose MLLMs \citep{hurst2024gpt4o, team2023gemini, liu2023llava1, liu2024llava2} and task-specific models \citep{cheng2024emotion, zhang2023emotionclip, binz2025foundation} have been developed. However, while general-purpose MLLMs demonstrate strong performance across a broad range of tasks, they often lack the fine-grained specialization required for nuanced affective and social understanding \citep{cheng2024emotion}; task-specific models address this gap within their target domains but exhibit limited generalizability beyond them. 

\noindent\textbf{User Simulation in Conversational AI.}
User simulation has been a long-standing approach for training and evaluating task-oriented dialogue systems without costly human involvement. In the pre-LLM era, early methods modeled user behavior through rule-based and probabilistic frameworks \citep{schatzmann2007agenda, schatzmann2009hidden, pietquin2006probabilistic}, while the rise of neural networks enabled data-driven approaches that progressively replaced hand-crafted rules by learning user behavior directly from corpora \citep{asri2016sequence, gur2018user, shi2019build}. In the LLM era, user simulators have grown substantially more naturalistic and flexible, evolving from generative transformer-based methods \citep{lin2022gentus, sun2023metaphorical} to more sophisticated LLM-based approaches that improve goal consistency, reduce hallucinations, and address goal alignment \citep{sekulic2024reliable, davidson2023user, mehri2025goal}. However, existing simulators are largely confined to text-only, task-oriented settings, leaving the rich non-verbal and affective dimensions of real human interaction unexplored. Benefiting from the rapid advancement of MLLMs, SayNext-Bench addresses this gap by simulating users' latent social-cognitive states—including but not limited to goal, intention, and emotion—from multimodal dialogue context. 

\vspace{-2mm}

\section{SayNext-Bench}

\vspace{-2mm}


We introduce \textsc{SayNext-Bench} to evaluate whether MLLMs can develop human-like anticipatory abilities in conversation. From a human-machine interaction perspective, this also serves as a probe into MLLMs' ability to simulate comprehensive social-cognitive user states, encompassing emotion, intention, and related cognitive dimensions, from multimodal input. Since such cognitive mechanisms are notoriously difficult to measure directly, we formulate next-utterance anticipation as a \textbf{proxy} task.

\textbf{Task Setups.}
Firstly, we define \textit{conversational turn} as a paired unit of discourse, consisting of an interlocutor’s utterance and a subject’s temporally contiguous response, which jointly serve as the fundamental granularity for evaluation. The \textsc{SayNext} task is then formulated as anticipating the forthcoming utterance $\tilde{T}_B^{\mathcal{R}}$ of the subject, conditioned on the interlocutor’s preceding utterance $T_A^{\mathcal{Q}}$ and the subject’s concurrent non-verbal expressions $V_B^{\mathcal{Q}}$. 

The anticipation problem can be expressed as approximating the conditional distribution of real responses $T_B^{\mathcal{R}}$ via an end-to-end mapping function $f_{\theta}$:
\vspace{-2.8mm}
\begin{equation}
\begin{aligned}
P\!\left(\tilde{T}_B^{\mathcal{R}} \mid T_A^{\mathcal{Q}}, V_B^{\mathcal{Q}}\right)
&\approx 
P\!\left(T_B^{\mathcal{R}} \mid T_A^{\mathcal{Q}}, V_B^{\mathcal{Q}}\right), \\
f_{\theta} &: (T_A^{\mathcal{Q}}, V_B^{\mathcal{Q}}) \mapsto \tilde{T}_B^{\mathcal{R}} .
\end{aligned}
\label{eq:mapping-function}
\end{equation}
\textbf{The SayNext-PC Dataset.}
Based on the task definition, we find that no existing dataset fully satisfies all requirements of our task. Specifically, an eligible dataset must meet three essential criteria:  
(1) dyadic interaction, involving either two humans or one human and one machine;  
(2) multimodality, with at least synchronized video and text modalities;  
(3) continuous viewpoint, where the camera remains steadily focused on one interlocutor throughout the interaction. Frequent camera shifts or interruptions impede the reliable capture of facial and bodily expressions that are critical for inferring communicative intent and emotion. A detailed dataset investigation is provided in Appendix~\ref{ap: dataset discussion}.  

To instantiate these principles, we introduce SayNext-PC2K, a novel dataset for next-utterance anticipation, inspired by the post-match press conference setting in the iMiGUE dataset~\citep{liu2021imigue}. For scalability, we further expand it to SayNext-PC19K, augmenting 359 original recordings to 3,463 videos from multiple tournaments. Both datasets capture athlete–interviewer dialogues with stable camera views, ensuring clear speech and expressive non-verbal cues. 
Finally, the SayNext-PC2K comprises 2,092 minutes of dialogue spanning 5,432 turns, while SayNext-PC19K contains 20,766 minutes with 38,540 turns. Each dialogue is segmented into video clips and transcribed using Whisper~\citep{whisper2023}. The transcription quality was verified, yielding an average Word Error Rate (WER) of 4.11\%, which is within the range generally considered human-acceptable (see Appendix~\ref{ap:wer-experiment} for details). We note that the constrained social context of post-match press conferences, with well-defined discourse norms and predictable emotional valence, reduces response stochasticity and provides a controlled, tractable setting for this inherently challenging task, serving as a principled starting point before scaling to more open-ended domains.

\textbf{Evaluation Metrics.}
Evaluating next-utterance anticipation is non-trivial due to its inherent one-to-many nature: even humans rarely produce word-for-word identical responses, making exact lexical matching (e.g. BLEU-4) an insufficient criterion. Such surface-form metrics fail to capture whether the underlying intent, emotion, and semantics are correctly anticipated. We therefore carefully designed a \textbf{three-level evaluation framework}: \textit{Lexical Similarity} (Level 1), \textit{Emotion-Intention Consistency} (Level 2), and \textit{Overall Alignment} (Level 3). This multi-level design aims to comprehensively capture the anticipation quality as follows:

\noindent\hspace{1em}\textit{\textbf{Level 1: Lexical Similarity.}}
Following standard NLP evaluation practice, this level provides a direct measurement of utterance-level similarity across two complementary dimensions. Lexical Overlap (LO) is measured by BLEU-4 \textbf{(LO-B)}~\citep{papineni-etal-2002-bleu} and ROUGE-L \textbf{(LO-R)}~\citep{lin-2004-rouge}, which quantify word- and phrase-level correspondence between anticipated and reference responses. Semantic Similarity (SS) is assessed by BERTScore-F1 \textbf{(SS-B)}~\citep{bertscore2020} and Sentence-BERT \textbf{(SS-S)}~\citep{reimers-gurevych-2019-sentence}, capturing token-level and sentence-level semantic alignment, respectively.

\noindent\hspace{1em}\textit{\textbf{Level 2: Emotion-Intention Consistency.}} Following the well-recognized circumplex model of affect~\citep{russell1980circumplex} and the taxonomy of empathetic response intents~\citep{welivita2020taxonomy}, this level evaluates whether a model preserves the affective state and communicative intention underlying the reference response. \textbf{Continuous Emotion Consistency (CEC)} is quantified by Valence \textbf{(CEC-V)} and Arousal \textbf{(CEC-A)} via normalized profile comparisons from the NRC-VAD lexicon~\citep{mohammad-2018-obtaining}. \textbf{Discrete Emotion Consistency (DEC)} and \textbf{Intention Consistency (IC)} are measured by sentence-level label agreement using dedicated classifiers covering 7 emotion~\citep{busso2008iemocap} and 9 intention categories~\citep{welivita2020taxonomy} respectively, with details in Appendix~\ref{ap: metrics}.

\noindent\hspace{1em}\textit{\textbf{Level 3: Overall Alignment.}}
This level employs LLM-as-Judge to provide a holistic, preference-based assessment of whether anticipations match the overall feeling of the reference response. GPT-4.1 \textbf{($J_{\text{GPT}}$)}  and Gemini-2.5-Pro \textbf{($J_{\text{Gem}}$)} serve as evaluators, ranking model anticipations by their alignment with the ground truth. 
All metric details and evaluation prompts are provided in Appendix~\ref{ap: metrics} and~\ref{ap:prompt}. Sec.~\ref{sec: human & LLM test} also reports user studies to provide human-aligned perspectives.

\textbf{Evaluation Protocol.} SayNext-Bench adopts four complementary evaluation protocols to assess model robustness across subject variability, generalization regimes, and data scales. Specifically, the \emph{subject-dependent} protocol allows training and test instances from the same subject to overlap, enabling models to capture individual-specific expressive patterns under a fixed 4:1 split, while the \emph{subject-independent} protocol enforces disjoint subject sets between training and testing to evaluate generalization to unseen speakers, with balanced representation across five continents. To examine cross-context robustness, the \emph{cross-scenario} protocol examines whether model mechanisms generalize across datasets with diverse conversational scenarios, allowing fine-tuning on target data to verify that performance gains are not tied to a single dataset. Finally, the \emph{scalability} protocol extends SayNext-PC2K to the larger SayNext-PC19K, testing model stability under increased temporal span and cultural diversity. Detailed protocol definitions and dataset construction procedures are provided in Appendix~\ref{ap: evaluation-protocol}.

\begin{figure*}[t]
  \centering
  \includegraphics[width=0.90\linewidth]{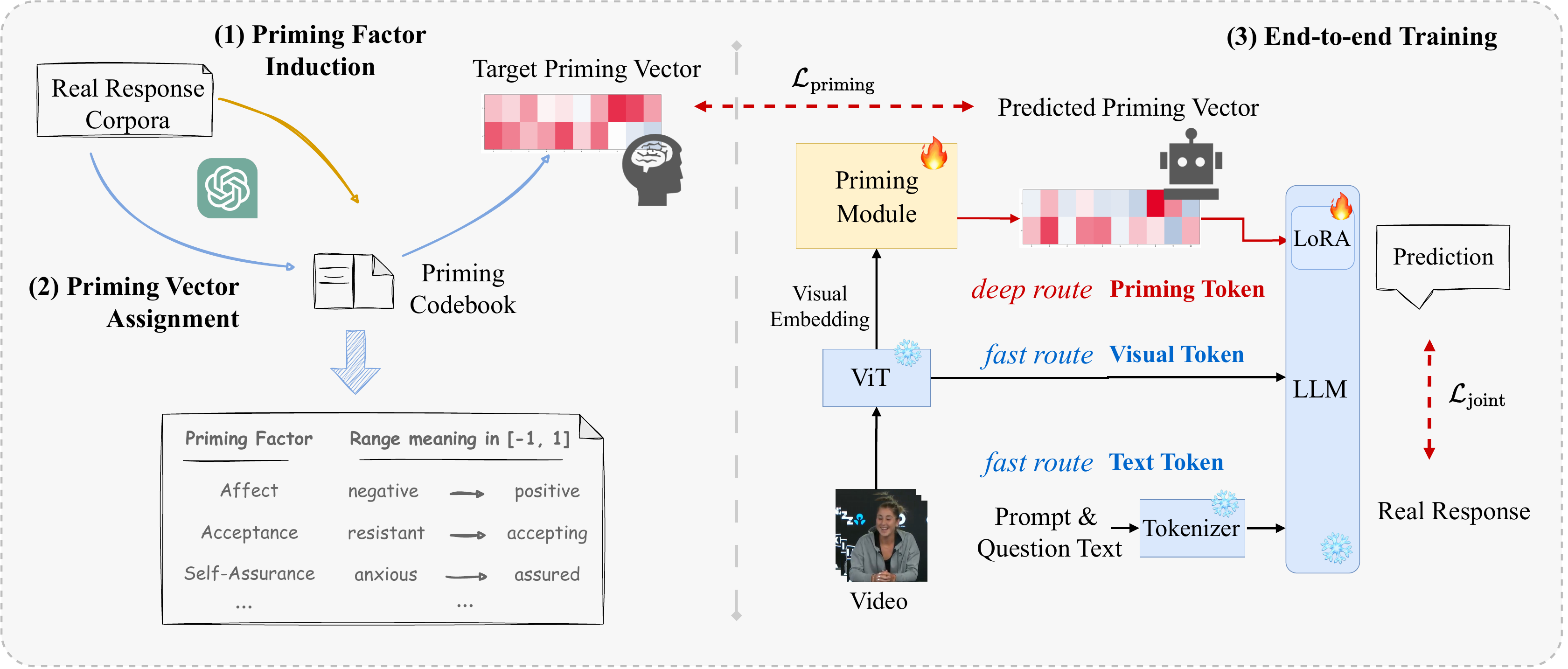}
    \caption{{\small \textbf{The SayNext-Chat Framework.} (1) Priming factors are extracted through LLM-assisted induction to construct a priming codebook. (\textcolor{myyellow}{yellow arrows}) (2) The codebook guides the LLM in assigning a target priming vector to each response. (\textcolor{myblue}{blue arrows}) (3) During end-to-end training, the loss combines the MSE between target and predicted priming vectors with the cross-entropy loss from the LLM backbone.}}
  \label{fig:flow}
  \vspace{-0.4cm}
\end{figure*}

\vspace{-0.2cm}
\section{The Proposed Method}

\vspace{-0.2cm}


We propose a dual-route prediction framework, SayNext-Chat, to anticipate forthcoming utterances. This framework incorporates learnable priming tokens that represent high-level belief priors from visual inputs and low-level cues directly perceived from multimodal inputs. An overview of the framework is shown in Figure~\ref{fig:flow}. We first introduce the generation of the priming factor and then describe its role within the dual-route framework.


\subsection{Priming factor}  
\label{sec:priming factor}


Rooted in the cognitive psychology concept of \textit{semantic priming}, whereby prior exposure to a stimulus facilitates processing of related information by pre-activating contextually relevant representations before speech~\citep{meyer1971facilitation, neely1977semantic, neely2012semantic, clark2013whatever}, we operationalize this notion computationally to capture predictive contextual priors over forthcoming utterances. Specifically, we leverage an LLM (GPT-4.1\footnote{\href{https://platform.openai.com/docs/models/gpt-4.1}{Model card: gpt-4.1-2025-04-14}}) to uncover recurrent, domain-relevant semantic associations from the training split only, which are then embedded and clustered into coherent latent structures. Each cluster is distilled by the LLM into a concise \textbf{priming factor}, constrained to be a neutral, widely recognized phrase denoting a specific behavioral or psychological characteristic. Along with the factor label, the LLM specifies a factor-specific polarity schema—\textbf{priming codebook}—that defines the meanings of its positive and negative poles. We then assign every response a \textbf{target priming vector}, corresponding to a set of priming factors generated by the LLM. The value of each entry in the priming vector lies in $[-1,1]$: the magnitude encodes activation strength, and the sign encodes factor-specific polarity ($+1$ = strong positive manifestation, $-1$ = strong negative manifestation, $0$ = not manifested). In this way, the target priming vector serves as an auxiliary representation that \textit{actively} encodes contextual prior beliefs about forthcoming utterances. To our knowledge, this is the \textbf{first attempt} to incorporate multimodal cues as priming factors into LLM frameworks for anticipatory dialogue modeling. The full priming codebook is shown in Table~\ref{tab:priming_codebook_2K}.

\vspace{-0.2cm}

\subsection{SayNext-Chat: Dual-route Prediction Framework}

Inspired by dual\mbox{-}route accounts of human cognition and the role of priors~\citep{kahneman2011thinking, clark2013whatever}, we design SayNext\mbox{-}Chat with two complementary predictive routes: a \textbf{fast route} that directly maps low-level visual and textual cues to a response and a \textbf{deep route} that infers high-level priors (priming factors) to guide generation. \textbf{Fast route.} A visual backbone InternViT\mbox{-}300M, an MLP projector, and an LLM InternLM2.5\mbox{-}7B conditioned on prompts, text, and visual embeddings to produce the next utterance. 
\textbf{Deep route.} Non\mbox{-}verbal features are transformed by two convolutional layers and a fully connected layer into a \emph{priming embedding}, injected into the LLM as a dedicated priming token (concatenated with visual tokens). In parallel, the embedding is mapped by another fully connected layer to a \emph{priming vector}, supervised by the target priming vector from Sec.~\ref{sec:priming factor}. Incorporating priming as learnable tokens improves output stability (Sec.~\ref{sec:priming-ablation}) and aligns with hierarchical representations in cognition.

\vspace{-0.2cm}

\textbf{Training Strategy.}
During training, both visual and language backbones are fine-tuned with Low-Rank Adaptation (LoRA)~\citep{hu2022lora} using rank $r=16$, with the resulting adapter weights merged back into the backbone after training. The priming module is trained from scratch. The training objective consists of two components: the language modeling loss $\mathcal{L}_{\mathrm{joint}}$, defined as the standard cross-entropy over tokens predicted and ground-truth sequences, and the priming loss $\mathcal{L}_{\mathrm{prim}}$, defined as a mean squared error regression loss encouraging the predicted priming vector to align with the target vector.

To balance optimization, we employ an \textit{adaptive loss weighting} scheme. 
The overall training loss is 
\begin{equation}
\mathcal{L} = \mathcal{L}_{\text{joint}} + \lambda_{\text{prim}} \, \mathcal{L}_{\text{prim}},
\end{equation}
where $\lambda_{\text{prim}}$ dynamically adjusts the relative importance of the priming loss. Specifically, it is updated according to an exponential moving average (EMA) of $\mathcal{L}_{\text{prim}}$: 
\begin{equation}
\begin{aligned}
\lambda_{\text{prim}}
&= \min \!\left(
    \max \!\left(
        \tfrac{\tilde{\mathcal{L}}_{\text{prim}}^{(t)}}{\mathcal{L}_{\text{prim}}^{(0)}+\epsilon},
        \, 0
    \right),
    \, 1
\right), \\
\tilde{\mathcal{L}}_{\text{prim}}^{(t)}
&= \mu \, \tilde{\mathcal{L}}_{\text{prim}}^{(t-1)}
  + (1-\mu) \, \mathcal{L}_{\text{prim}}^{(t)} .
\end{aligned}
\end{equation}
Here, $\tilde{\mathcal{L}}_{\text{prim}}^{(t)}$ denotes the smoothed priming loss at step $t$. We set $\mu=0.99$ and $\epsilon$ as a small constant for numerical stability. The training regimen employs AdamW optimizer with cosine learning rate decay initialized at $1\times10^{-4}$, and takes 2$\times$A100 GPUs for training.

\vspace{-2mm}

\section{Experiments}
\subsection{Comparison with State-of-the-art Models}
\label{sec: main result section}
We evaluate SayNext-Chat against seven state-of-the-art MLLM baselines, grouped into three categories:  
(1) frontier large-scale MLLMs, including GPT4o\footnote{\href{https://platform.openai.com/docs/models/gpt-4o}{Model card: gpt-4o}} and Gemini2.5-flash\footnote{\href{https://storage.googleapis.com/deepmind-media/Model-Cards/Gemini-2-5-Flash-Model-Card.pdf}{Model card: Gemini 2.5 Flash}};  
(2) open-source MLLMs with comparable parameter scales (7–8B) for fair comparison, including InternVL2-8B~\citep{chen2024internvl}, VideoLLaMA3-7B~\citep{zhang2025videollama}, LLaVA-NeXT-Video-7B~\citep{zhang2024llavanext-video}, and InstructBLIP-7B~\citep{dai2023instructblip};  
and (3) task-specialized MLLMs for emotion recognition, represented by Emotion-LLaMA-7B~\mbox{\citep{cheng2024emotion}}.  
All baselines are evaluated under identical prompts with a temperature of 0.7. Notably, given the computational cost of LLM-based evaluation (Level 3) and the participant burden of user studies, we restrict these assessments to the primary subject-dependent protocol, where differences across models are most pronounced. Implementation details are in Appendix~\ref{ap: experiment detail}.

Figure~\ref{fig:main-res} presents the main results under the subject-dependent protocol. (a) reports the average rank of all models, showing that SayNext-Chat combines relatively small size with superior performance. (b) highlights notable discrepancies in Level 1 and Level 2 evaluations: lexical overlap remains extremely low (below 20\%), semantic similarity reaches a moderate level (around 0.4--0.6), and emotion-intention consistency achieves the highest scores (around 0.7--0.9). These results underscore the inherent difficulty of reproducing word-level utterances, even with advanced fine-tuned MLLMs. To further illustrate model distinctions, (c) shows that \textbf{SayNext-Chat consistently surpasses all state-of-the-art baselines across every metric}, substantially improving emotion-intention consistency by capturing latent connections between non-verbal cues and their linguistic realizations.

The no-vision baseline (gray lines in Figure~\ref{fig:main-res}(c)) consistently falls below SayNext-Chat across all Level 1 and Level 2 metrics, confirming that visual cues contribute meaningfully to anticipating users' social-cognitive states. Notably, it still surpasses several vision-equipped baselines on DEC and IC, suggesting textual context alone carries strong priors for discrete emotion and intention classification. Attention visualizations in Figure~\ref{fig:sample-test}(b) and Appendix~\ref{ap: attn-vis} further show that SayNext-Chat attends more broadly to facial expressions and gesture regions. Performance trends remain consistent across both protocols, though several metrics decrease slightly in the subject-independent setting, reflecting the challenge of modeling subject-specific patterns (Appendix~\ref{ap: tables}).

Figure~\ref{fig:sample-test} (a) presents case studies comparing priming-vector heatmaps with ground-truth and generated responses. High-score cases typically arise when the question provides sufficient context, salient non-verbal expressions enable affective alignment, and the dialogue follows common patterns (e.g., joy after victory, pressure from an opponent). In contrast, low-score cases often involve off-topic noise and the inability of current MLLMs to reproduce human humor or linguistic creativity. More cases shown in Appendix~\ref{ap:case study-1}.

\begin{figure*}[t]
  \centering
  \includegraphics[width=\linewidth]{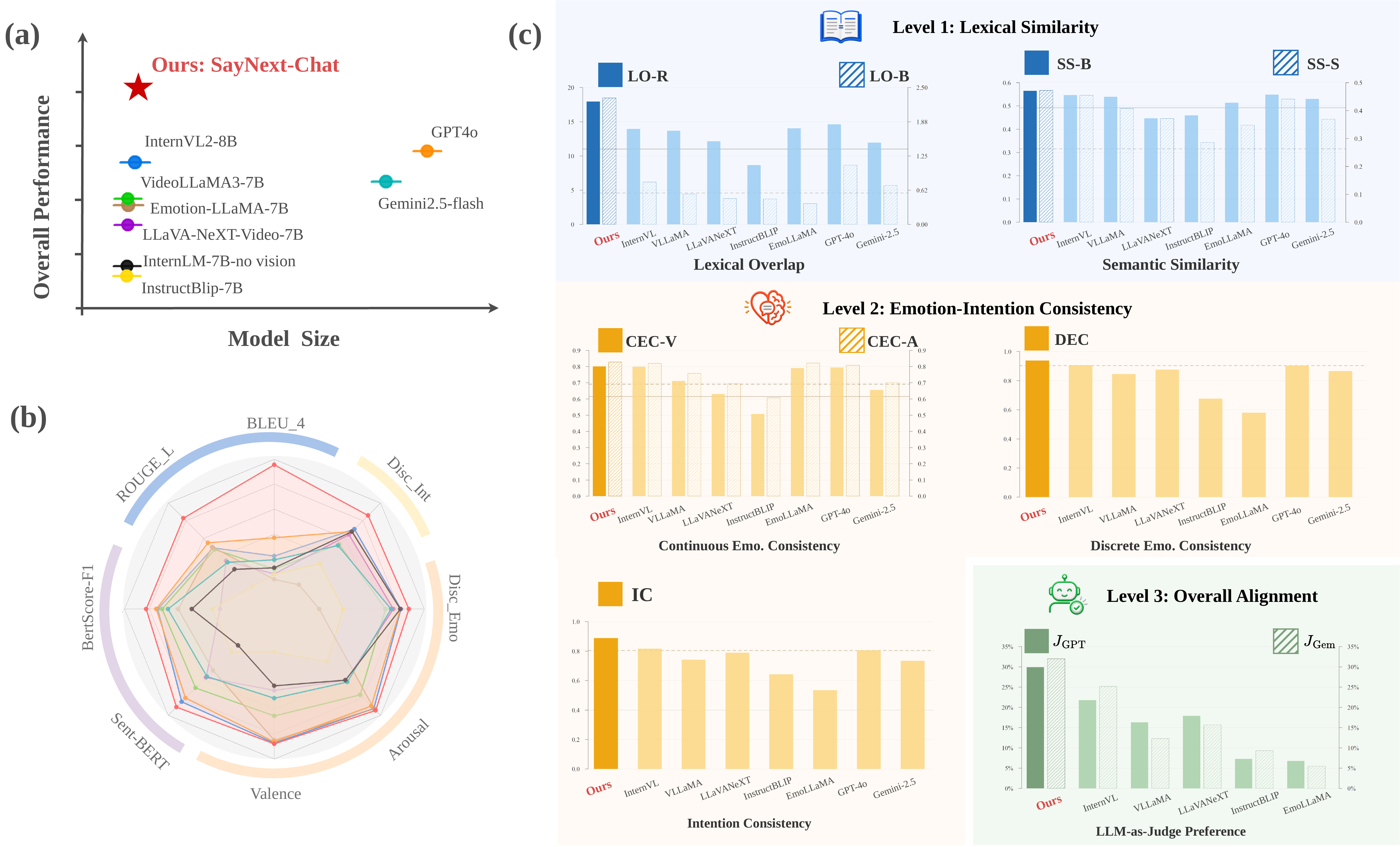}
    \caption{\textbf{Comparison of SayNext-Chat with State-of-the-art Baselines.} (a) Two-dimensional comparison of relative model size and performance (average rank across metrics). SayNext-Chat (red star) achieves both smaller size and higher performance. (b) Radar chart of multi-metric evaluation, where each polygon corresponds to a model and is colored consistently with (a). Our model (red) consistently ranks highest across all metrics in the radar chart. (c) Bar charts of ten metrics across three levels. SayNext-Chat (the first bar) consistently attains the highest score. Solid and dashed horizontal lines indicate the no-vision baseline for filled and hatched bars, respectively.}
  \label{fig:main-res}
  \vspace{-0.6cm}
\end{figure*}

\vspace{-3mm}

\subsection{Ablation Study of the Priming Token}
\label{sec:priming-ablation}

\vspace{-2mm}

To assess the contribution of priming information, we conduct ablation studies across two baseline models used in Sec. \ref{sec: main result section}, comparing three variants:
(1) a fine-tuned model without the priming module,
(2) prompt-based variants that inject priming factors or target priming vectors via prompt engineering, and
(3) SayNext-Chat, which incorporates the priming vector as a dedicated learnable token.
All ablations are performed on the SayNext-PC2K dataset under the subject-dependent protocol.

\begin{table}[!t]
\centering
\caption{
Relative deltas (higher is better) w.r.t.\ different backbones.
Priming generally improves most metrics, especially emotion-intention consistency, across backbones.
While fine-tuning alone may degrade emotion alignment.
}
\label{tab:ablation_prim_all}
\resizebox{\textwidth}{!}{%
\begin{tabular}{cc cccc cccc}
\toprule
\multirow{3}{*}{Backbone} & \multirow{3}{*}{Method}
  & \multicolumn{4}{c}{\textbf{Level 1: Lexical Similarity}}
  & \multicolumn{4}{c}{\textbf{Level 2: Emotion-Intention Consistency}} \\
\cmidrule(lr){3-6} \cmidrule(lr){7-10}
  & & \multicolumn{2}{c}{\makecell{Lexical\\Overlap /\%}}
  & \multicolumn{2}{c}{\makecell{Semantic\\Sim.}}
  & \multicolumn{2}{c}{\makecell{Cont. Emo.\\Consist.}}
  & \makecell{Dis. Emo.\\Consist.}
  & \makecell{Intent.\\Consist.} \\
\cmidrule(lr){3-4} \cmidrule(lr){5-6} \cmidrule(lr){7-8}
  & & $\Delta$LO-B$\uparrow$ & $\Delta$LO-R$\uparrow$
  & $\Delta$SS-B$\uparrow$ & $\Delta$SS-S$\uparrow$
  & $\Delta$CEC-V$\uparrow$ & $\Delta$CEC-A$\uparrow$
  & $\Delta$DEC$\uparrow$ & $\Delta$IC$\uparrow$ \\
\midrule

\multirow{4}{*}{\textit{InternVL2-8B}}
& Factor in Prompt
& $-0.708$ & $-3.343$ & $-0.03988$ & $-0.07260$ & $-0.10859$ & $-0.08465$ & $-0.09979$ & $-0.09082$ \\
& Vector in Prompt
& $-0.715$ & $-3.278$ & $-0.03517$ & $-0.06980$ & $-0.08949$ & $-0.08249$ & $-0.09642$ & $-0.12409$ \\
& Only Finetune
& $+1.308$ & $+3.664$ & \best{+0.01870} & \best{+0.01770} & $-0.02506$ & $-0.01751$ & $+0.00768$ & $+0.03758$ \\
& Finetune+Priming (Ours)
& \cellcolor[HTML]{F0F0F0}{\best{+1.535}}
& \cellcolor[HTML]{F0F0F0}{\best{+4.021}}
& \cellcolor[HTML]{F0F0F0}{$+0.01830$}
& \cellcolor[HTML]{F0F0F0}{$+0.01760$}
& \cellcolor[HTML]{F0F0F0}{\best{+0.00277}}
& \cellcolor[HTML]{F0F0F0}{\best{+0.00878}}
& \cellcolor[HTML]{F0F0F0}{\best{+0.03139}}
& \cellcolor[HTML]{F0F0F0}{\best{+0.07017}} \\

\midrule

\multirow{4}{*}{\textit{VideoLLaMA3-7B}}
& Factor in Prompt
& $-0.425$ & $-2.189$ & $-0.01173$ & $-0.04620$ & $-0.01678$ & $-0.02802$ & $-0.00916$ & $-0.02984$ \\
& Vector in Prompt
& $-0.445$ & $-1.931$ & $-0.00570$ & $-0.05310$ & $-0.02608$ & $-0.04135$ & $-0.02683$ & $-0.02981$ \\
& Only Finetune
& $+1.309$ & $+3.586$ & $+0.02332$ & $+0.04560$ & $+0.06546$ & $+0.04309$ & $+0.06166$ & $+0.10080$ \\
& Finetune+Priming (Ours)
& \cellcolor[HTML]{F0F0F0}{\best{+1.558}}
& \cellcolor[HTML]{F0F0F0}{\best{+3.816}}
& \cellcolor[HTML]{F0F0F0}{\best{+0.02501}}
& \cellcolor[HTML]{F0F0F0}{\best{+0.05930}}
& \cellcolor[HTML]{F0F0F0}{\best{+0.07069}}
& \cellcolor[HTML]{F0F0F0}{\best{+0.04652}}
& \cellcolor[HTML]{F0F0F0}{\best{+0.07647}}
& \cellcolor[HTML]{F0F0F0}{\best{+0.12450}} \\

\bottomrule
\end{tabular}%
}
\vspace{-0.3cm}
\end{table}
\begin{table}[t]
\centering
\begin{minipage}[t]{0.495\textwidth}
\centering
\caption{Cross-scenario Evaluation Results on IEMOCAP.}
\label{tab:cross-scen}
\huge
\setlength{\tabcolsep}{2.5pt}
\resizebox{\linewidth}{!}{
\begin{tabular}{l cccc cccc}
\toprule
\multirow{3}{*}[-3pt]{Model}
  & \multicolumn{4}{c}{\textbf{Lv.1: Lexical Sim.}}
  & \multicolumn{4}{c}{\textbf{Lv.2: Emo.-Int. Consist.}} \\
\cmidrule(lr){2-5} \cmidrule(lr){6-9}
  & \multicolumn{2}{c}{\makecell{Lexical\\Overlap /\%}}
  & \multicolumn{2}{c}{\makecell{Semantic\\Sim.}}
  & \multicolumn{2}{c}{\makecell{Cont. Emo.\\Consist.}}
  & \makecell{Dis. Emo.\\Consist.}
  & \makecell{Intent.\\Consist.} \\
\cmidrule(lr){2-3} \cmidrule(lr){4-5} \cmidrule(lr){6-7}
  & LO-B$\uparrow$ & LO-R$\uparrow$ & SS-B$\uparrow$ & SS-S$\uparrow$ & CEC-V$\uparrow$ & CEC-A$\uparrow$ & DEC$\uparrow$ & IC$\uparrow$ \\
\midrule
InternVL2   & 0.44 & 9.60 & 0.50 & 0.18 & 0.51 & 0.57 & 0.64 & 0.43 \\
VideoLLaMA3    & 0.38 & 9.53 & 0.49 & 0.17 & 0.45 & 0.51 & 0.50 & 0.32 \\
LLaVA-NeXT     & 0.44 & 8.21 & 0.49 & 0.15 & 0.39 & 0.47 & 0.59 & 0.30 \\
Emotion-LLaMA  & 0.26 & 10.06 & 0.45 & 0.17 & 0.53 & \textbf{0.62} & 0.31 & 0.28 \\
GPT-4o         & 0.47 & 9.50 & 0.47 & 0.18 & 0.44 & 0.50 & 0.52 & 0.40 \\
Gemini-2.5     & 0.91 & 9.13 & 0.50 & 0.15 & 0.44 & 0.50 & 0.53 & 0.36 \\
\cellcolor[HTML]{F0F0F0}{\textbf{SayNext-Chat}} & \cellcolor[HTML]{F0F0F0}{\textbf{5.44}} & \cellcolor[HTML]{F0F0F0}{\textbf{22.29}} & \cellcolor[HTML]{F0F0F0}{\textbf{0.58}} & \cellcolor[HTML]{F0F0F0}{\textbf{0.31}} & \cellcolor[HTML]{F0F0F0}{\textbf{0.55}} & \cellcolor[HTML]{F0F0F0}{0.59} & \cellcolor[HTML]{F0F0F0}{\textbf{0.69}} & \cellcolor[HTML]{F0F0F0}{\textbf{0.54}} \\
\bottomrule
\end{tabular}
}
\end{minipage}\hfill
\begin{minipage}[t]{0.495\textwidth}
\centering
\caption{Scalability Evaluation Results on SayNext-PC19K.}
\label{tab:scale-res}
\huge
\setlength{\tabcolsep}{2.5pt}
\vspace{1ex}
\resizebox{\linewidth}{!}{
\begin{tabular}{l cccc cccc}
\toprule
\multirow{3}{*}[-3pt]{Model}
  & \multicolumn{4}{c}{\textbf{Lv.1: Lexical Sim.}}
  & \multicolumn{4}{c}{\textbf{Lv.2: Emo.-Int. Consist.}} \\
\cmidrule(lr){2-5} \cmidrule(lr){6-9}
  & \multicolumn{2}{c}{\makecell{Lexical\\Overlap /\%}}
  & \multicolumn{2}{c}{\makecell{Semantic\\Sim.}}
  & \multicolumn{2}{c}{\makecell{Cont. Emo.\\Consist.}}
  & \makecell{Dis. Emo.\\Consist.}
  & \makecell{Intent.\\Consist.} \\
\cmidrule(lr){2-3} \cmidrule(lr){4-5} \cmidrule(lr){6-7}
  & LO-B$\uparrow$ & LO-R$\uparrow$ & SS-B$\uparrow$ & SS-S$\uparrow$ & CEC-V$\uparrow$ & CEC-A$\uparrow$ & DEC$\uparrow$ & IC$\uparrow$ \\
\midrule
InternVL2    & 0.36 & 13.04 & 0.54 & 0.47 & \textbf{0.79} & 0.81 & 0.92 & 0.85 \\
VideoLLaMA3     & 0.18 & 11.91 & 0.52 & 0.41 & 0.73 & 0.77 & 0.85 & 0.78 \\
LLaVA-NeXT      & 0.46 & 13.79 & 0.53 & 0.46 & 0.79 & 0.80 & 0.92 & 0.85 \\
Emotion-LLaMA   & 0.25 & 12.66 & 0.50 & 0.36 & 0.79 & 0.81 & 0.63 & 0.60 \\
\cellcolor[HTML]{F0F0F0}{\textbf{SayNext-Chat}} & \cellcolor[HTML]{F0F0F0}{\textbf{2.49}} & \cellcolor[HTML]{F0F0F0}{\textbf{15.71}} & \cellcolor[HTML]{F0F0F0}{\textbf{0.55}} & \cellcolor[HTML]{F0F0F0}{\textbf{0.47}} & \cellcolor[HTML]{F0F0F0}{0.79} & \cellcolor[HTML]{F0F0F0}{\textbf{0.82}} & \cellcolor[HTML]{F0F0F0}{\textbf{0.94}} & \cellcolor[HTML]{F0F0F0}{\textbf{0.90}} \\
\bottomrule
\end{tabular}
}
\end{minipage}
\end{table}

\begin{table}[t]
\centering
\begin{minipage}[t]{0.495\textwidth}
\centering
\caption{User study results. SayNext-Chat achieves the highest best-selection rate across all three experimental groups. \(E\langle N\rangle\) denotes the experimental group number.}
\label{tab:userstudy1}
\small
\setlength{\tabcolsep}{4pt}
\vspace{3ex}
\resizebox{\linewidth}{!}{
    \begin{tabular}{lcccc}
    \toprule
    Model & E1 & E2 & E3 & Average \\
    \midrule
    InternVL2  & 24.25\% & 25.50\% & 24.00\% & 24.58\% \\
    VideoLLaMA3      & 17.50\% & 13.50\% & 11.75\% & 14.25\% \\
    GPT-4o   & 26.00\% & 26.75\% & 28.00\% & 26.92\% \\  
    \cellcolor[HTML]{F0F0F0}{\textbf{SayNext-Chat}}    & \cellcolor[HTML]{F0F0F0}{\textbf{32.25\%}} & \cellcolor[HTML]{F0F0F0}{\textbf{34.25\%}} & \cellcolor[HTML]{F0F0F0}{\textbf{36.25\%}} & \cellcolor[HTML]{F0F0F0}{\textbf{34.25\%}} \\
    \bottomrule
    \end{tabular}
}
\end{minipage}
\hfill
\begin{minipage}[t]{0.495\textwidth}
\caption{LLM evaluation results (Level 3). SayNext-Chat attains the highest top-1 rate under both GPT-4.1 and Gemini2.5-Pro rankings.}
\label{tab:LLM-study}
\small
\setlength{\tabcolsep}{4pt}
\vspace{1ex}
\resizebox{\linewidth}{!}{
    \begin{tabular}{lccc}
    \toprule
    Model & GPT4.1 &Gemini2.5-pro  & Average \\
    \midrule
    InternVL2  & 21.78\% & 25.19\%  & 23.48\% \\
    VideoLLaMA3      & 16.30\% & 12.30\%  & 14.30\% \\
    Emotion-LLaMA  & 6.81\% & 5.48\%  & 6.15\% \\
    LLaVA-NeXT      & 17.93\% & 15.70\%  & 16.81\% \\
    InstructBLIP      & 7.26\% & 9.33\%  & 8.30\% \\
   \cellcolor[HTML]{F0F0F0}{\textbf{SayNext-Chat}}    & \cellcolor[HTML]{F0F0F0}{\textbf{29.93\%}} & \cellcolor[HTML]{F0F0F0}{\textbf{32.00\%}} & \cellcolor[HTML]{F0F0F0}{\textbf{30.96\%}} \\
    \bottomrule
    \end{tabular}
}
\end{minipage}%
\hfill
\end{table}

As shown in Table~\ref{tab:ablation_prim_all}, fine-tuning alone enables the model to acquire domain-relevant latent representations, while incorporating priming vectors further enhances sensitivity to salient semantic cues, yielding steady gains in emotion-intention consistency over the fine-tuned baseline.
Results on VideoLLaMA3 show that all metrics consistently benefit from the introduction of the priming module, further indicating that our method improves model performance beyond a single backbone.
In contrast, prompt-based approaches perform poorly, as explicit priming factors or vectors tend to bias the model toward surface-level numerical cues rather than supporting the anticipation task.
Overall, these findings highlight that \textbf{the stable integration of cognitive priors through learnable tokens is indispensable for modeling anticipatory dialogue}.

\vspace{-3mm}

\subsection{Cross-scenario Generalization and Scalability}
\label{sec:cross-scale and domain}

For cross-scenario evaluation, we adapt IEMOCAP to SayNext-Bench. As shown in Table~\ref{tab:cross-scen}, SayNext-Chat consistently outperforms all state-of-the-art zero-shot baselines, with only the CEC-A dimension slightly below EmotionLlama—an expected outcome given EmotionLlama’s explicit emphasis on emotion. These results demonstrate \textbf{the benchmark’s extensibility across scenarios and the model’s capacity for cross-dataset generalization}. For scalability, as shown in Table~\ref{tab:scale-res}, SayNext-Chat also achieves the best overall performance on most metrics in the larger SayNext-PC19K dataset, despite its broader cultural diversity and extended temporal span. The overall metric distributions remain comparable to those observed on SayNext-PC2K, with only a slight decrease in CEC-V.

Additionally, a comparison of extracted priming factors across the three datasets reveals both scenario-specific patterns and stable recurring dimensions, such as \textit{Affect}, \textit{Confidence}, and \textit{Disappointment}. This convergence suggests that certain cognitive–affective factors are broadly universal across conversational contexts. More broadly, it implies that scaling the size and diversity of datasets may support the development of a generalized inventory of priming vectors, potentially applicable to everyday dialogue and informative for cognitive modeling.

\begin{figure}[!t]
  \centering
  \includegraphics[width=\linewidth]{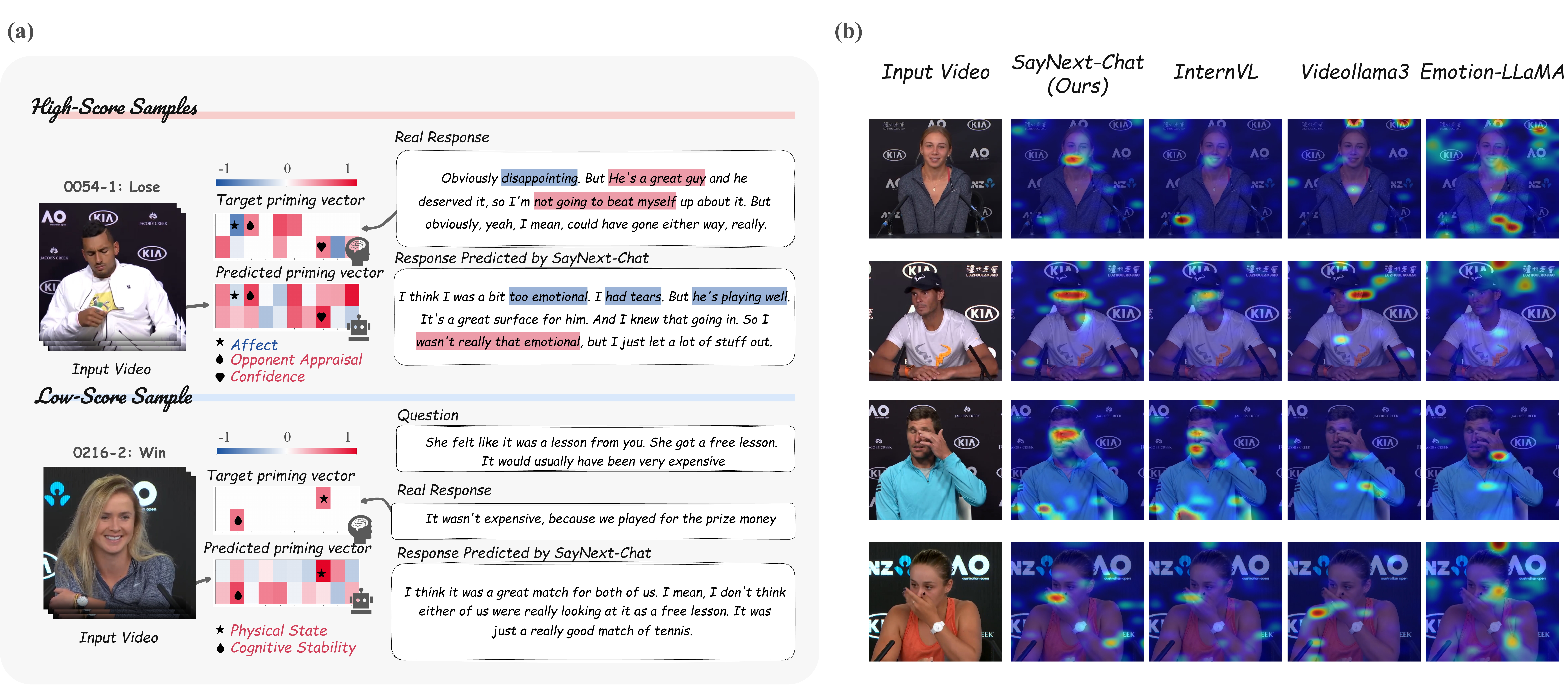}
  \vspace{0.5ex}
\caption{\textbf{Case Study on SayNext-PC2K.} 
(a) In high-score samples, the predicted priming vector heatmap closely 
matches the target. \colorbox{lightred}{Red} and \colorbox{lightblue}{Blue} 
indicate positive and negative values, with corresponding highlights in the 
response text. Star, heart, and drop markers denote three representative 
priming factors and their semantic meaning (listed beneath the heatmap). 
Low-score samples exhibit target priming patterns that are difficult to predict.
(b) SayNext-Chat attends more broadly to facial expressions (rows 1--2) and 
gesture regions (rows 3--4) compared to baselines. 
Full attention visualizations are in Appendix~\ref{ap: attn-vis}.}
  \label{fig:sample-test}
  \vspace{-0.4cm}
\end{figure}

\subsection{User Study \& LLM-as-Judge Evaluation}
\label{sec: human & LLM test}
Beyond standard metrics, we conduct two subjective evaluations. In the user study, 12 participants (3 groups) evaluated the same 400-question test set, selecting the candidate most similar to the reference from four models; outputs were randomized to mitigate order effects (Table~\ref{tab:userstudy1}). For LLM-as-Judge, GPT-4.1 and Gemini2.5-Pro\footnote{\href{https://storage.googleapis.com/model-cards/documents/gemini-2.5-pro.pdf}{Model card: Gemini-2.5-Pro}} ranked anticipations from five baselines and our model, excluding GPT-4o and Gemini to avoid RLHF-induced family bias (Table~\ref{tab:LLM-study}).
SayNext-Chat achieves the highest user-study selection rate (34.25\%) and ranks first under both LLM evaluators, with consistent preference distributions across groups and evaluators, suggesting high inter-evaluator agreement.


\vspace{-2mm}
\section{Limitations \& Future Works}
\label{sec:dis-and-limi}


Although results on SayNext-Bench indicate the potential of LLMs to anticipate forthcoming utterances from multimodal cues, notable limitations remain. Low lexical-overlap accuracy reflects the intrinsic difficulty of this task and highlights the need for more cognitively capable MLLMs. As discussed in Appendix~\ref{ap: multiturn}, individual variability and the stochastic nature of language further motivate finer-grained modeling in multi-turn settings. In addition, dataset quality remains a constraint, as noisy samples with modality inconsistencies or off-topic content can hinder effective learning. Ethical considerations are discussed in Appendix~\ref{ap:ethics}.


\textbf{Future directions}. Extending \textsc{SayNext} to more nuanced pragmatic–emotional expressions, such as sarcasm, humor, and metaphor, remains challenging even for advanced MLLMs but is a natural next step. Integrating multi-turn or richer multimodal context may further help capture speaker-specific habits and discourse dynamics. Moreover, developing cognitively grounded evaluation frameworks that better capture intention alignment and anticipatory reasoning, potentially through collaboration with cognitive science and social psychology, is essential. Finally, analyzing the learned priming factors as cognitively meaningful dimensions offers an opportunity to build more interpretable and human-aligned anticipatory models.
\vspace{-2mm}
\section{Conclusion}
\label{sec:conclusion}
\vspace{-2mm}

We propose \textsc{SayNext}, a multimodal next-utterance anticipation task for trustworthy, seamless human–AI dialogue, and release SayNext-Bench with a carefully designed multi-level evaluation framework, the scalable SayNext-PC dataset, and four complementary evaluation protocols. Inspired by cognitive neuroscience, we develop SayNext-Chat, a dual-route MLLM that injects learnable priming tokens to fuse perceptual cues with anticipatory priors. Across subject-dependent/independent settings and cross-scenario datasets, SayNext-Chat consistently surpasses strong baselines across all evaluation levels, corroborated by user studies and LLM-as-Judge evaluations. Ablation studies confirm that priming tokens and visual information both materially improve anticipation quality. Together, these findings validate the feasibility and potential of \textsc{SayNext}, offering a step toward bridging cognitively grounded mechanisms of human communication with the design of more empathetic, adaptive, and trustworthy interactive AI agents.

\newpage
\bibliographystyle{unsrtnat}
\bibliography{example_paper}


\newpage

\appendix

\newpage
\section*{\centering Appendix Table of Contents}
\vspace{1em}
\startcontents[appendices]
\printcontents[appendices]{l}{1}{\setcounter{tocdepth}{3}}
\newpage

\section{Technical appendices and supplementary material}

\subsection{LLM Usage Statement}
\label{ap:llm-usage}

Large Language Models (ChatGPT) were used exclusively to improve the clarity and fluency of English writing. They were not involved in research ideation, experimental design, data analysis, or interpretation. The authors take full responsibility for all content.

\subsection{Dataset}
\label{ap: dataset discussion}

\subsubsection{Dataset Comparison}
Although there are plenty of existing multimodal emotion recognition datasets (in Table~\ref{tab:dataset_comparison}), we find they all fail to be directly applied to our proposed task. For example, the widely used MELD dataset \citep{poria2018meld} — composed of video segments cropped from television shows — suffers from frequent viewpoint changes, which impede the reliable capture of visual expressions of subjects for emotion inference. Recent datasets such as MaSaC \citep{bedi2021masac} and CPED \citep{chen2022cped} exhibit similar issues, which we refer to as the “\textit{Interrupted}” view in this paper. Although earlier datasets like IEMOCAP \citep{busso2008iemocap} and MSP-IMPROV \citep{busso2016msp} offer consistent recording of subjects in conversational videos (we refer as “\textit{Continuous}” recording view), they are limited by small sample sizes and a restricted number of speakers. 
The SEMAINE dataset \citep{mckeown2011semaine}, which records interactions between humans and virtual visual characters, is the closest match; however, due to language generation technology constraints in 2012, the verbal output from the virtual characters is ineffective, diminishing the meaning to anticipate natural language expression.
To fill this gap and inspired by \citep{liu2021imigue}, we introduce the \textit{SayNext-PC} dataset, specifically designed for the \textit{SayNext} benchmark. A comparative overview of \textit{SayNext-PC}, scale in PC2K and PC19K, and other related datasets is presented in Table \ref{tab:dataset_comparison}.

\begin{table}[h]
  \centering
    \caption{\textbf{Attributes comparison between SayNext-PC and other widely used datasets for multimodal emotion recognition in dialogue.} Note: \textit{\#} denotes the number of instances; \textit{(s/min)/(U/D)} indicates the average duration in seconds per utterance or per dialogue; \textit{F/M} represents female/male; \textit{A}: audio, \textit{V}: video, \textit{T}: text, \textit{M}: dynamic motion capture, \textit{Mi}: micro body expression.}
    \label{tab:dataset_comparison}
  \renewcommand{\arraystretch}{1.0}
    \newcommand{\cmark}{{\textbf{\textcolor{ForestGreen}{\ding{51}}}}}
    \newcommand{\xmark}{{\textbf{\textcolor{BrickRed}{\ding{55}}}}}
    {
    \scriptsize
  \begin{tabularx}{\linewidth}{lllllllcc}
    \hline
    \rule{0pt}{10pt} Dataset & \#Dia.  & \#Utter.  & Ave. Dur. & \#Subj. (F/M) & Res. & Modality & \makecell{Descriptive\\label} & \makecell{Continuous\\view} \\
    \hline
    \rule{0pt}{10pt} IEMOCAP & 5 & 10039 & 4.5s/U & 10 (5/5) & $720\times480$ & A/V/T/M & \xmark & \cmark \\
    \rule{0pt}{10pt} SEMAINE & 959 & 5816 & 5min/D & 150 & $720\times580$ & A/V/T & \xmark & \cmark \\
    \rule{0pt}{10pt} MSP-IMPROV & 6 & 7818 & 4s/U & 12 (6/6) & $1440\times1080$ & A/V & \xmark & \cmark \\
    \rule{0pt}{10pt} MELD & 1433 & 13708 & 3.59s/U & 260+ & $1280\times720$ & A/V/T & \xmark & \xmark \\
    \rule{0pt}{10pt} MaSaC & 1190 & 15576 & 20s/U & - & - & A/V/T & \xmark & \xmark \\
    \rule{0pt}{10pt} CPED & 12000 & 132762 & 2.1s/U & 392 & - & A/V/T & \xmark & \xmark \\
    \hline
    \rule{0pt}{10pt} \bf{\textit{SayNext-PC2K}} & \textbf{359} & \textbf{5432} & \textbf{22.44s/U} & \textbf{72 (36/36)} & {\boldmath $1280\times720$} & \textbf{A/V/T/Mi} & \cmark & \cmark \\
    \rule{0pt}{10pt} \bf{\textit{SayNext-PC19K}} & \textbf{3463} & \textbf{38540} & \textbf{30.368s/U} & \textbf{474 (246/228)} & {\boldmath $640\times360$} & \textbf{A/V/T} & \cmark & \cmark \\
    \hline
  \end{tabularx}
  }
\end{table}

\subsubsection{Dataset Settings}

\vspace{-0.2cm}

\textbf{The SayNext-PC2K dataset} comprising 359 post-match press conference videos from Grand Slam tournaments was collected from online platforms, yielding a total duration of 2,092 minutes. All videos were recorded at a resolution of \(1280\times720\) and a frame rate of 25 fps. Micro-body language annotations were directly adopted as non-verbal visual labels. Speaker diarization was employed to segment the videos into question-answer pairs, resulting in 2716 pairs with an average utterance duration of 22.44 seconds.

Based on two evaluation protocols discussed in the manuscript, two data split settings were employed: subject-dependent and subject-independent. 

In the subject-dependent setting, samples were randomly assigned to the training and test sets, with careful partitioning to ensure that each subject appears in both sets. To achieve a 4:1 ratio, the training dataset comprises 2,041 samples, while the test dataset consists of 675 samples.

In the subject-independent evaluation, a comprehensive statistical analysis was performed to examine the distribution of subject identifiers and their corresponding nationalities. The training set comprises 58 unique subject IDs, representing a diverse spectrum of nationalities including:
\begin{itemize}
    \item \textit{Spain, Bulgaria, Russia, the United Kingdom, Switzerland, France, Germany, Belgium, Austria, the United States, Croatia, Canada, the Czech Republic, Australia, Denmark, Slovakia, South Korea, South Africa, Serbia, Argentina, Romania, Japan, Taiwan, Latvia, Greece, China, Italy.}
\end{itemize}
In contrast, the test set contains 14 unique subject IDs, with nationalities including:
 \begin{itemize}
    \item \textit{Switzerland, Japan, the United Kingdom, Canada, Australia, Germany, Spain, Russia, Ukraine, the United States, China, South Africa.}
\end{itemize}
Notably, the test set spans five continents, thereby underscoring the cultural universality of the dataset. These statistics highlight the international diversity within both the training and test subsets, which in turn attests to the robustness and representativeness of our data.
To achieve a 4:1 ratio, the training dataset comprises 2,155 samples, while the test dataset consists of 561 samples.

\begin{figure*}[!t]
  \centering
  \includegraphics[width=\linewidth]{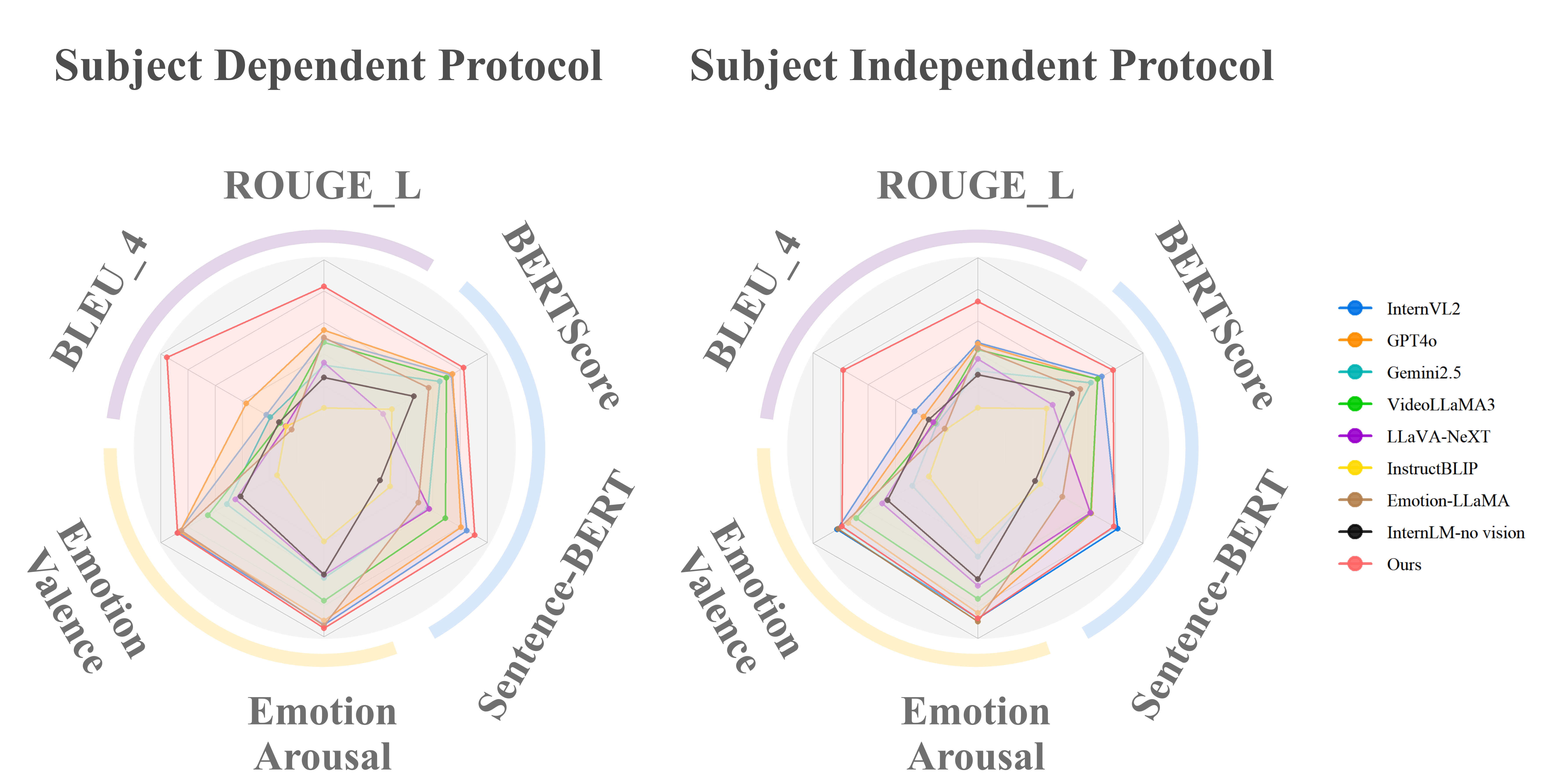}
  \caption{\textbf{Multidimensional comparison between baseline models and our method in subject-dependent and subject-independent protocols.} Our model (red) forms the largest polygon in both subject-dependent and subject-independent settings, highlighting its strong overall performance across both protocols.}
  \label{fig:radar}
\end{figure*}

\textbf{The SayNext-PC19K dataset} comprises 3,463 post-match press conference videos collected from a wider spectrum of tournaments, including the Australian Open, US Open, and Wimbledon. Relative to SayNext-PC2K (2017–2019), the temporal coverage has been substantially extended to span 2017–2024, thereby encompassing more recent competitions and reflecting the evolving dynamics of professional tennis discourse. The number of participating athletes has also expanded markedly, from 72 to 474, which introduces cross-cultural characteristics and heterogeneous communicative norms and expressive tendencies. In terms of data quality, one quarter of the videos are available at a resolution of \(1280\times720\), while the remainder are at \(640\times360\). Each video is meticulously annotated with a unique athlete identifier and nationality, ensuring traceability across sessions and enabling systematic analysis of subject-level and cultural factors.

\textbf{The IEMOCAP dataset} is an interactive emotional dyadic motion-capture corpus collected by the Speech Analysis \& Interpretation Laboratory at the University of Southern California. It consists of dyadic sessions in which 10 actors (5 female and 5 male) perform either improvisations or scripted scenarios, yielding approximately 12 hours of audiovisual data, including video, speech, facial motion capture, and text transcriptions. The sessions are manually segmented into utterances, each annotated by at least three human annotators. For our \textsc{SayNext} task, we adopt the dataset based on its transcriptions, which have been shown to exhibit a low Word Error Rate (WER). Nevertheless, the utterances are much shorter in duration compared with SayNext-PC, reflecting the context-sparse nature of the IEMOCAP corpus and making next-utterance anticipation in single-turn conversations particularly challenging.

To provide a clearer overview of its scenario diversity, we classify and enumerate the major dialogue scenarios in IEMOCAP in Table~\ref{tab:iemocap_scenario}. 

\begin{table}[h]
\centering
\small
\caption{Coarse-to-fine categorization of IEMOCAP dialogue scenarios}
\label{tab:iemocap_scenario}
\resizebox{\textwidth}{!}{%
\begin{tabularx}{\textwidth}{p{3cm} p{7cm} p{3cm}}
\toprule
\textbf{Coarse Category} & \textbf{Fine-grained Scenario} & \textbf{Appears in} \\
\midrule

\textbf{Workplace / Administrative} &
Customer--service conflict (e.g., ID office argument) &
Improvised \\
& Bureaucratic frustration (forms, documents, procedures) &
Improvised \\
& Clerk--customer misunderstanding &
Improvised \\
& Workplace tension / poor service attitude &
Improvised \\
\midrule

\textbf{Casual / Light Conversation} &
Neutral conversation before conflict escalation &
Scripted \\
& Small-talk transitions embedded in scripted scenes &
Scripted \\
& Everyday conversational prelude &
Minimal presence \\
\midrule

\textbf{Family / Domestic} &
Family dispute about a missing or presumed-dead relative &
Scripted \\
& Parent--child argument about responsibilities &
Scripted \\
& Sibling disagreement / family stress &
Scripted \& Improvised \\
& Coping with grief inside the family &
Scripted \\
\midrule

\textbf{Romantic / Interpersonal} &
Couple jealousy argument &
Improvised \\
& Break-up confrontation &
Improvised \\
& Suspicion or betrayal between partners &
Improvised \\
& Reconciliation attempt &
Improvised \\
\midrule

\textbf{Supportive / Apology} &
Apologizing after doing something wrong &
Improvised \\
& Consoling a grieving friend &
Improvised \\
& Comforting someone after a negative event &
Improvised \\
& Emotional reassurance / empathetic support &
Improvised \\
\midrule

\textbf{General Conflict} &
Accusation and denial &
Improvised \\
& Heated disagreement between acquaintances &
Scripted \& Improvised \\
& Blaming someone for wrongdoing &
Improvised \\
& Confronting someone about past actions &
Scripted \\
\midrule

\parbox[t]{4cm}{\textbf{Miscellaneous}\\\textbf{(Scripted Drama)}} &
Classic theatre conflict (e.g., ``Who's Afraid of Virginia Woolf?'', ``The Apartment'') &
Scripted \\
& Emotional monologues inside dramatic scenes &
Scripted \\
& High-tension theatrical exchanges &
Scripted \\
\bottomrule
\end{tabularx}
}%
\end{table}

\subsection{Experiment Details}
\label{ap: experiment detail}
\subsubsection{Evaluation Protocols Details}
\label{ap: evaluation-protocol}
\noindent\textbf{Subject-Dependent Evaluation.} (Results in Sec. \ref{sec: main result section}) In this setting, instances from the same subject in SayNext-PC2K appear in both the training and test sets. Conversational turns are randomly partitioned such that each subject in the test set has at least one corresponding instance in the training set, enabling the model to capture individual-specific nuances in emotional and behavioral expression. The split follows a 4:1 ratio between training and test data and is fixed the same for all models. 

\noindent\textbf{Subject-Independent Evaluation.} (Results in Sec. \ref{sec: main result section})
In this setting, SayNext-PC2K is split such that the subjects in the training and test sets form disjoint groups; that is, all subjects in the test set are unseen during training. The dataset comprises 72 subjects, divided into training and test groups at a 4:1 ratio. To mitigate potential cultural bias, subjects from five continents (North America, Australia, Europe, Asia, and Africa) are evenly represented across both groups.

\noindent\textbf{Cross-Scenario Evaluation.} (Results in Sec. \ref{sec:cross-scale and domain})
To examine robustness across conversational contexts, we adapt the IEMOCAP dataset~\citep{busso2008iemocap}, a dyadic motion-capture corpus comprising 12 hours of scripted and improvised dyadic dialogues. The dataset spans diverse scenarios such as job interviews, relationship conflicts, and casual exchanges, from which 4,113 conversational turns are extracted for our task. This protocol is intended to verify whether trained models can \textit{directly} generalize from the solely press-conference scenario to different conversational settings in a zero-shot manner.

\noindent\textbf{Scalability Evaluation.} (Results in Sec. \ref{sec:cross-scale and domain})  
To assess the robustness of model performance at large scales, we construct SayNext-PC19K, a substantially larger dataset that extends the temporal span of conversations and incorporates broader cultural diversity. This broader scope is specifically designed to avoid potential overfitting issues with local optima in SayNext-PC2K, thereby enhancing the capacity of SayNext-Bench to evaluate models under more diverse and fine-grained conditions.

\subsubsection{Experimental Setups}

\textbf{Video Input.}  Videos are processed as sequences of frames, with up to 4 frames used during fine-tuning and up to 16 frames during inference. Frames are extracted at intervals of 8. Specifically, in fine-tuning, 4 frames are randomly sampled in each epoch using a sliding-window strategy, which provides dynamic visual information while reducing memory costs. For consistency, all models employing InternViT as the visual backbone (zero-shot baselines, fine-tuned models, and \textsc{SayNext}) resize frames to \(480\times480\).

\textbf{Inference Configuration.}
For zero-shot baselines, we evaluate the \textsc{SayNext} task using seven state-of-the-art MLLMs. Owing to the inherent complexity of these models, numerous hyperparameters and configuration settings must be considered during fine-tuning. To clarify our experimental setup, we highlight several important modifications. In InstructBLIP, only a single frame is used for anticipation, as the model lacks a dedicated video modality input. Similarly, in EmotionLLaMA, we also utilize a single frame, since our experiments indicate that this configuration yields better performance. This improvement is likely attributable to the reduced redundancy in visual embeddings that typically arises when processing multiple frames. VideoLLaMA3 and LLaVA-NeXT, by contrast, are provided with four input frames, consistent with our model. For GPT-4o and Gemini-2.5-Flash, we extract four frames for GPT-4o and adapt Gemini’s native video interface to process visual information at approximately 1 fps.

For all baselines and our model, we apply the same prompt to generate anticipations (full prompt in Appendix~\ref{ap:prompt}). For InternLM-based inference, we adopt consistent generation settings as follows:

\begin{minipage}{\columnwidth}
\begin{verbatim}
    no_repeat_ngram_size=2,
    repetition_penalty=1,  
    do_sample=True,               
    temperature=0.7,             
    top_k=50,                   
    top_p=0.9,             
    max_new_tokens=2048
\end{verbatim}
\end{minipage}

For all fine-tuned variants, we apply the same LoRA configuration to the visual and language backbones unless otherwise specified.

\subsubsection{Full Result Tables}
\label{ap: tables}

We present the complete experimental results (tables and figures) in this section, providing numerical evidence for accurate reference.

First, Figure~\ref{fig:radar} shows the radar plots of the subject-dependent and subject-independent protocols. Both exhibit similar hexagonal patterns, where our model achieves the best overall performance. Across baselines, the no-vision model (black) forms the smallest hexagon, while GPT-4o and InternVL2 perform better than the remaining baselines. Notably, the hexagon of our model under the subject-independent protocol is slightly smaller than that under the subject-dependent setting, indicating that subject-independent evaluation is more challenging—likely due to the model’s limited ability to capture individual-specific patterns.

The detailed quantitative results for both subject-dependent and subject-independent protocols are reported in the following two tables (Tables~\ref{tab:full-1} and~\ref{tab:full-2}).

Table~\ref{tab:full-3} reports the experimental results on IEMOCAP, demonstrating the transferability of our model across different contexts and scenarios.

Table~\ref{tab:full-4} presents the experimental results on SayNext-PC19K, demonstrating the generalizability of our model to larger-scale datasets. The results further suggest potential scale effects in the \textsc{SayNext} task and benchmark, indicating that training on sufficiently large and fine-grained corpora could foster the development of more generalizable next-utterance anticipation models. Such models would lay the groundwork for more advanced conversational AI and for trustworthy, safe human–computer interaction, while also offering an AI-driven perspective on the mechanisms underlying human cognition.

\begin{table*}[!t]
    \centering
    \caption{\textbf{Multi-level numerical results on SayNext-PC2K under Subject-Dependent Protocol.}
    Best in \textbf{bold}, second-best \underline{underlined}.
    BLEU-4 and ROUGE-L are in percentages (\%).
    $^\dagger$GPT-4o and Gemini-2.5 are excluded from Level~3 evaluation
    as they serve as the LLM judges themselves. $^\dagger$The no-vision baseline is excluded from Level 3 evaluation as it serves as a lower-bound reference rather than a competitive system.}
    \label{tab:full-1}
    \setlength{\belowcaptionskip}{-4mm}
    \footnotesize
    \setlength{\tabcolsep}{4pt}
    \renewcommand{\arraystretch}{1.3}
    \begin{tabular}{l cccc cccc cc}
    \hline
    \multirow{3}{*}{\textbf{Method}}
      & \multicolumn{4}{c}{\textbf{Level 1: Lexical Similarity}}
      & \multicolumn{4}{c}{\textbf{Level 2: Emotion-Intention Consistency}}
      & \multicolumn{2}{c}{\textbf{Level 3: Overall}} \\
    \cmidrule(lr){2-5} \cmidrule(lr){6-9} \cmidrule(lr){10-11}
      & \multicolumn{2}{c}{\makecell{Lexical \\Overlap /\%}}
      & \multicolumn{2}{c}{\makecell{Semantic \\Sim.}}
      & \multicolumn{2}{c}{\makecell{Cont. Emo.\\Consist.}}
      & \makecell{Dis. Emo.\\Consist.}
      & \makecell{Intent.\\Consist.}
      & \multicolumn{2}{c}{\makecell{LLM-as-Judge\\Pref.}} \\
    \cmidrule(lr){2-3} \cmidrule(lr){4-5} \cmidrule(lr){6-7} \cmidrule(lr){10-11}
      & LO-B$\uparrow$ & LO-R$\uparrow$
      & SS-B$\uparrow$ & SS-S$\uparrow$
      & CEC-V$\uparrow$ & CEC-A$\uparrow$
      & DEC$\uparrow$
      & IC$\uparrow$
      & $J_{\text{GPT}}$$\uparrow$ & $J_{\text{Gem}}$$\uparrow$ \\
    \hline
        No vision      & 0.574 & 11.005 & 0.4921 & 0.2632 & 0.61549 & 0.69160 & 0.9037 & 0.8044 &  $^\dagger$--- &  $^\dagger$--- \\
        InternVL2      & 0.772 & 13.936 & 0.5468 & \underline{0.4546} & \underline{0.79863} & 0.81969 & \underline{0.9064} & \underline{0.8172} & \underline{21.78\%} & \underline{25.19\%} \\
        VideoLLaMA3    & 0.554 & 13.696 & 0.5397 & 0.4072 & 0.71166 & 0.75820 & 0.8450 & 0.7407 & 16.30\% & 12.30\% \\
        LLaVA-NeXT     & 0.472 & 12.143 & 0.4469 & 0.3715 & 0.63044 & 0.69315 & 0.8755 & 0.7894 & 17.93\% & 15.70\% \\
        InstructBlip   & 0.466 & 8.679  & 0.4596 & 0.2856 & 0.50761 & 0.60748 & 0.6758 & 0.6424 & 7.26\% & 9.33\% \\
        Emotion-LLaMA  & 0.380 & 14.046 & 0.5139 & 0.3478 & 0.78967 & \underline{0.82126} & 0.5793 & 0.5348 & 6.81\% & 5.48\% \\
        GPT-4o         & \underline{1.081} & \underline{14.623} & \underline{0.5489} & 0.4415 & 0.79441 & 0.80838 & 0.9034 & 0.8053 & $^\dagger$--- & $^\dagger$--- \\
        Gemini-2.5     & 0.710 & 11.951 & 0.5300 & 0.3686 & 0.65538 & 0.70022 & 0.8663 & 0.7340 & $^\dagger$--- & $^\dagger$--- \\
    \hline
        SayNext-Chat & \textbf{2.307} & \textbf{17.957} & \textbf{0.5651} & \textbf{0.4722} & \textbf{0.80140} & \textbf{0.82847} & \textbf{0.9378} & \textbf{0.8874} & \textbf{29.93\%} & \textbf{32.00\%} \\
    \hline
    \end{tabular}
\end{table*}

\begin{table*}[!t]
    \centering
    \caption{\textbf{Numerical results on SayNext-PC2K on Subject-Independent Evaluation Protocol.} Best results shown in \textbf{bold}, and the second-best is underlined. The proposed approach achieves superior performance across all metrics. Note: BLEU-4 and ROUGE-L values are in percentages.}
    \label{tab:full-2}
    \setlength{\belowcaptionskip}{-4mm}
    \footnotesize
    \setlength{\tabcolsep}{4pt}
    \renewcommand{\arraystretch}{1.3}
    \begin{tabular}{l cccc cccc}
    \hline
    \multirow{3}{*}{\textbf{Method}}
      & \multicolumn{4}{c}{\textbf{Level 1: Lexical Similarity}}
      & \multicolumn{4}{c}{\textbf{Level 2: Emotion-Intention Consistency}} \\
    \cmidrule(lr){2-5} \cmidrule(lr){6-9}
      & \multicolumn{2}{c}{\makecell{Lexical\\Overlap /\%}}
      & \multicolumn{2}{c}{\makecell{Semantic\\Sim.}}
      & \multicolumn{2}{c}{\makecell{Cont. Emo.\\Consist.}}
      & \makecell{Dis. Emo.\\Consist.}
      & \makecell{Intent.\\Consist.} \\
    \cmidrule(lr){2-3} \cmidrule(lr){4-5} \cmidrule(lr){6-7}
      & LO-B$\uparrow$ & LO-R$\uparrow$ & SS-B$\uparrow$ & SS-S$\uparrow$ & CEC-V$\uparrow$ & CEC-A$\uparrow$ & DEC$\uparrow$ & IC$\uparrow$ \\
    \hline
        No vision    & 0.633 & 11.148 & 0.4968 & 0.2644 & 0.63313 & 0.70029 & \underline{0.9305} & 0.8164 \\
        InternVL2    & \underline{0.850} & \underline{13.576} & \underline{0.5402} & \textbf{0.4447} & \textbf{0.78024} & 0.79932 & 0.9234 & 0.8503 \\
        VideoLLaMA3  & 0.534 & 13.033 & 0.5339 & 0.3857 & 0.72401 & 0.75002 & 0.8582 & 0.7487 \\
        LLaVA-NeXT  & 0.565 & 12.332 & 0.4686 & 0.3852 & 0.64883 & 0.71677 & 0.8577 & 0.7971 \\
        InstructBlip & 0.371 & 8.643 & 0.4596 & 0.2759 & 0.51285 & 0.60562 & 0.6465 & 0.6062 \\
        Emotion-LLaMA  & 0.389 & 13.144 & 0.5086 & 0.3240 & \underline{0.77705} & \textbf{0.80732} & 0.5526 & 0.5437 \\
        GPT4o        & 0.710 & 13.470 & 0.5334 & 0.3872 & 0.74770 & 0.78584 & \textbf{0.9499} & \underline{0.8605} \\
        Gemini2.5       & 0.510 & 11.446 & 0.5244 & 0.2731 & 0.56094 & 0.64392 & 0.9250 & 0.7357 \\
        \hline
        SayNext-Chat \textbf{(Ours)} & \textbf{1.939} & \textbf{16.681} & \textbf{0.5563} & \underline{0.4365} & 0.76580 & \underline{0.79932} & 0.9198 & \textbf{0.8734} \\
    \hline
    \end{tabular}
\end{table*}
\begin{table*}[!t]
    \centering
    \caption{\textbf{Numerical results on Cross-Scenarios Evaluation Protocol.} The train dataset and test dataset are both from adapted IEMOCAP dataset. Best results shown in \textbf{bold}, and the second-best is underlined. The proposed approach achieves superior performance across all metrics. Note: BLEU-4 and ROUGE-L values are in percentages.}
    \label{tab:full-3}
    \setlength{\belowcaptionskip}{-4mm}
    \footnotesize
    \setlength{\tabcolsep}{4pt}
    \renewcommand{\arraystretch}{1.3}
    \begin{tabular}{l cccc cccc}
    \hline
    \multirow{3}{*}{\textbf{Method}}
      & \multicolumn{4}{c}{\textbf{Level 1: Lexical Similarity}}
      & \multicolumn{4}{c}{\textbf{Level 2: Emotion-Intention Consistency}} \\
    \cmidrule(lr){2-5} \cmidrule(lr){6-9}
      & \multicolumn{2}{c}{\makecell{Lexical\\Overlap /\%}}
      & \multicolumn{2}{c}{\makecell{Semantic\\Sim.}}
      & \multicolumn{2}{c}{\makecell{Cont. Emo.\\Consist.}}
      & \makecell{Dis. Emo.\\Consist.}
      & \makecell{Intent.\\Consist.} \\
    \cmidrule(lr){2-3} \cmidrule(lr){4-5} \cmidrule(lr){6-7}
      & LO-B$\uparrow$ & LO-R$\uparrow$ & SS-B$\uparrow$ & SS-S$\uparrow$ & CEC-V$\uparrow$ & CEC-A$\uparrow$ & DEC$\uparrow$ & IC$\uparrow$ \\
    \hline
        InternVL2    & 0.438 & 9.602 & \underline{0.5016} & \underline{0.1804} & 0.51430 & 0.57000 & \underline{0.6369} & \underline{0.4325} \\
        VideoLLaMA3  & 0.379 & 9.528 & 0.4915 & 0.1674 & 0.44701 & 0.51264 & 0.4987 & 0.3181 \\
        LLaVA-NeXT  & 0.164 & 7.047 & 0.4779 & 0.1724 & 0.39481 & 0.46541 & 0.5917 & 0.3038 \\
        Emotion-LLaMA  & 0.257 & \underline{10.058} & 0.4478 & 0.1724 & \underline{0.52532} & \textbf{0.61817} & 0.3054 & 0.2822 \\
        GPT4o        & 0.474 & 9.504 & 0.4707 & 0.1787 & 0.43655 & 0.49732 & 0.5186 & 0.3979 \\
        Gemini2.5       & \underline{0.914} & 9.126 & 0.4957 & 0.1528 & 0.43701 & 0.50057 & 0.5298 & 0.3593 \\
        \hline
        SayNext-Chat \textbf{(Ours)} & \textbf{5.439} & \textbf{22.292} & \textbf{0.57625} & \textbf{0.3076} & \textbf{0.54767} & \underline{0.59014} & \textbf{0.6902} & \textbf{0.5431} \\
    \hline
    \end{tabular}
\end{table*}
\begin{table*}[!t]
    \centering
    \caption{\textbf{Experimental results on SayNext-PC19K across different models.} Best results shown in \textbf{bold}, and the second-best is underlined. The proposed approach achieves superior performance across all metrics. Commercial models are not tested here due to high expense. Note: BLEU-4 and ROUGE-L values are in percentages.}
    \label{tab:full-4}
    \setlength{\belowcaptionskip}{-4mm}
    \footnotesize
    \setlength{\tabcolsep}{4pt}
    \renewcommand{\arraystretch}{1.3}
    \begin{tabular}{l cccc cccc}
    \hline
    \multirow{3}{*}{\textbf{Method}}
      & \multicolumn{4}{c}{\textbf{Level 1: Lexical Similarity}}
      & \multicolumn{4}{c}{\textbf{Level 2: Emotion-Intention Consistency}} \\
    \cmidrule(lr){2-5} \cmidrule(lr){6-9}
      & \multicolumn{2}{c}{\makecell{Lexical\\Overlap /\%}}
      & \multicolumn{2}{c}{\makecell{Semantic\\Sim.}}
      & \multicolumn{2}{c}{\makecell{Cont. Emo.\\Consist.}}
      & \makecell{Dis. Emo.\\Consist.}
      & \makecell{Intent.\\Consist.} \\
    \cmidrule(lr){2-3} \cmidrule(lr){4-5} \cmidrule(lr){6-7}
      & LO-B$\uparrow$ & LO-R$\uparrow$ & SS-B$\uparrow$ & SS-S$\uparrow$ & CEC-V$\uparrow$ & CEC-A$\uparrow$ & DEC$\uparrow$ & IC$\uparrow$ \\
    \hline
        InternVL2    & 0.358 & 13.042 & \underline{0.5387} & \underline{0.4679} & \textbf{0.79117} & 0.81085 & \underline{0.9215} & 0.8499 \\
        VideoLLaMA3  & 0.183 & 11.913 & 0.5181 & 0.4149 & 0.73393 & 0.76607 & 0.8514 & 0.7830 \\
        LLaVA-NeXT  & \underline{0.455} & \underline{13.792} & 0.5290  & 0.4593 & 0.78622 & 0.80323 & 0.9160 & \underline{0.8513} \\
        Emotion-LLaMA  & 0.254 & 12.664 & 0.5023 & 0.3558 & \underline{0.78894} & \underline{0.81350} & 0.6264 & 0.5983 \\
        \hline
        SayNext-Chat \textbf{(Ours)} & \textbf{2.493} & \textbf{15.709} & \textbf{0.5482} & \textbf{0.4712} & 0.78623 & \textbf{0.81864} & \textbf{0.9426} & \textbf{0.8970} \\
    \hline
    \end{tabular}
\end{table*}

\subsubsection{Additional Cross-Dataset Validation}
\label{app:cross_dataset}

To further verify that the observed performance trends are not due to adaptation artifacts, we perform additional cross-dataset validation under multiple training--testing combinations. Concretely, we consider: (i) training on our 2K subset and testing on IEMOCAP / 19K, (ii) training on IEMOCAP and testing on 2K / 19K, and (iii) training on 19K and testing on IEMOCAP. For each setting, we report all six evaluation metrics used in the main paper, covering lexical overlap, semantic similarity, and emotion consistency.

As summarized in Table~\ref{tab:cross_dataset}, models trained on larger and more diverse datasets (e.g., 19K) consistently achieve stronger generalization across test sets, particularly in semantic similarity (BERTScore-F1 and Sentence-BERT) and emotion consistency (Valence, Arousal). In contrast, models trained on smaller datasets (2K or IEMOCAP alone) exhibit notable drops when transferred to more challenging test distributions. Overall, these results reinforce that the performance patterns follow dataset size and difficulty, rather than being driven by any specific adaptation or overfitting artifact.

\subsubsection{Prompt-level Multi-turn Anticipation}
\label{ap: multiturn}

As discussed in the Limitations section, both individual variability and the inherently stochastic nature of language suggest that incorporating multi-turn dialogue context may better capture speaker-specific habits and background knowledge, thereby improving lexical alignment. To preliminarily assess whether SayNext-Chat can support longer-term dialogue modeling, we conduct a prompt-level multi-turn next-utterance anticipation experiment.

Specifically, we augment the input prompt with the preceding two rounds of dialogue and require the model to anticipate the response in the third conversational turn. To ensure a fair evaluation, the multi-turn dataset is carefully curated to prevent any data leakage between training and test splits. As shown in Table~\ref{tab:multi-turn}, the multi-turn setting consistently outperforms the single-turn baseline. These results indicate not only that SayNext-PC is applicable to both single-turn and multi-turn scenarios, but also that SayNext-Chat exhibits robust and improved performance when richer contextual history is available.

More importantly, and in line with our expectations, incorporating additional conversational context enhances the model’s ability to anticipate forthcoming human utterances. Nevertheless, more fine-grained, model-level approaches to multi-turn utterance anticipation remain an open research direction and are required to support personalization and adaptation in continuously evolving AI agents.

\begin{table*}[!t]
  \centering
  \footnotesize
  \setlength{\tabcolsep}{5pt}
  \caption{Cross-dataset validation under different training--testing combinations.}
  \label{tab:cross_dataset}
  \begin{tabular}{lcccccc}
    \toprule
      \multicolumn{1}{c}{Method} 
      & \multicolumn{2}{c}{\textbf{Lexical Overlap /\%}} 
      & \multicolumn{2}{c}{\textbf{Semantic Similarity}} 
      & \multicolumn{2}{c}{\textbf{Emotion Consistency}} \\
    \cmidrule(lr){1-1}
    \cmidrule(lr){2-3} \cmidrule(lr){4-5} \cmidrule(lr){6-7}
      Training $\rightarrow$ Testing
      & LO-B$\uparrow$ 
      & LO-R$\uparrow$ 
      & SS-B$\uparrow$ 
      & SS-S$\uparrow$ 
      & CEC-V$\uparrow$ 
      & CEC-A$\uparrow$ \\
    \midrule
    2K $\rightarrow$ IEMOCAP 
      & 0.893 & 10.695 & 0.4925 & 0.1382 & 0.55709 & 0.62108 \\
    2K $\rightarrow$ 19K 
      & 0.861 & 15.218 & 0.5475 & 0.4367 & 0.75789 & 0.78664 \\
    \midrule
    IEMOCAP $\rightarrow$ 19K 
      & 0.023 & 7.829 & 0.4644 & 0.1640 & 0.49756 & 0.60204 \\
    IEMOCAP $\rightarrow$ 2K 
      & 0.128 & 8.913 & 0.4747 & 0.1518 & 0.54655 & 0.63653 \\
    \midrule
    19K $\rightarrow$  IEMOCAP 
      & 1.156 & 10.901 & 0.5077 & 0.1491 & 0.51188 & 0.56456 \\
    19K $\rightarrow$ 2K 
      & 1.931 & 17.439 & 0.5655 & 0.4582 & 0.76968 & 0.79182 \\
    \bottomrule
  \end{tabular}
\end{table*}

\begin{table*}[!t]
\centering
\footnotesize
\caption{Comparison between single-turn and multi-turn next-utterance anticipation with SayNext-Chat.}
\label{tab:multi-turn}
\begin{tabular}{lcccccc}
\toprule
Model &    LO-B$\uparrow$ 
          & LO-R$\uparrow$ 
          & SS-B$\uparrow$ 
          & SS-S$\uparrow$ 
          & CEC-V$\uparrow$ 
          & CEC-A$\uparrow$ \\
\midrule
SayNext-Chat (Single-turn)& 2.307 & 17.957 & 0.5651 & 0.4722  & 0.80140 & 0.82847 \\
\cellcolor[HTML]{F0F0F0}{\textbf{SayNext-Chat (Multi-turn)}} & \cellcolor[HTML]{F0F0F0}{2.762} & \cellcolor[HTML]{F0F0F0}{18.176} & \cellcolor[HTML]{F0F0F0}{0.5685} & \cellcolor[HTML]{F0F0F0}{0.4887} & \cellcolor[HTML]{F0F0F0}{0.80441} & \cellcolor[HTML]{F0F0F0}{0.82786} \\
\bottomrule
\end{tabular}
\end{table*}

\subsubsection{Attention Visualizations}
\label{ap: attn-vis}
\begin{figure}[!t]
  \centering
  \includegraphics[width=\linewidth]{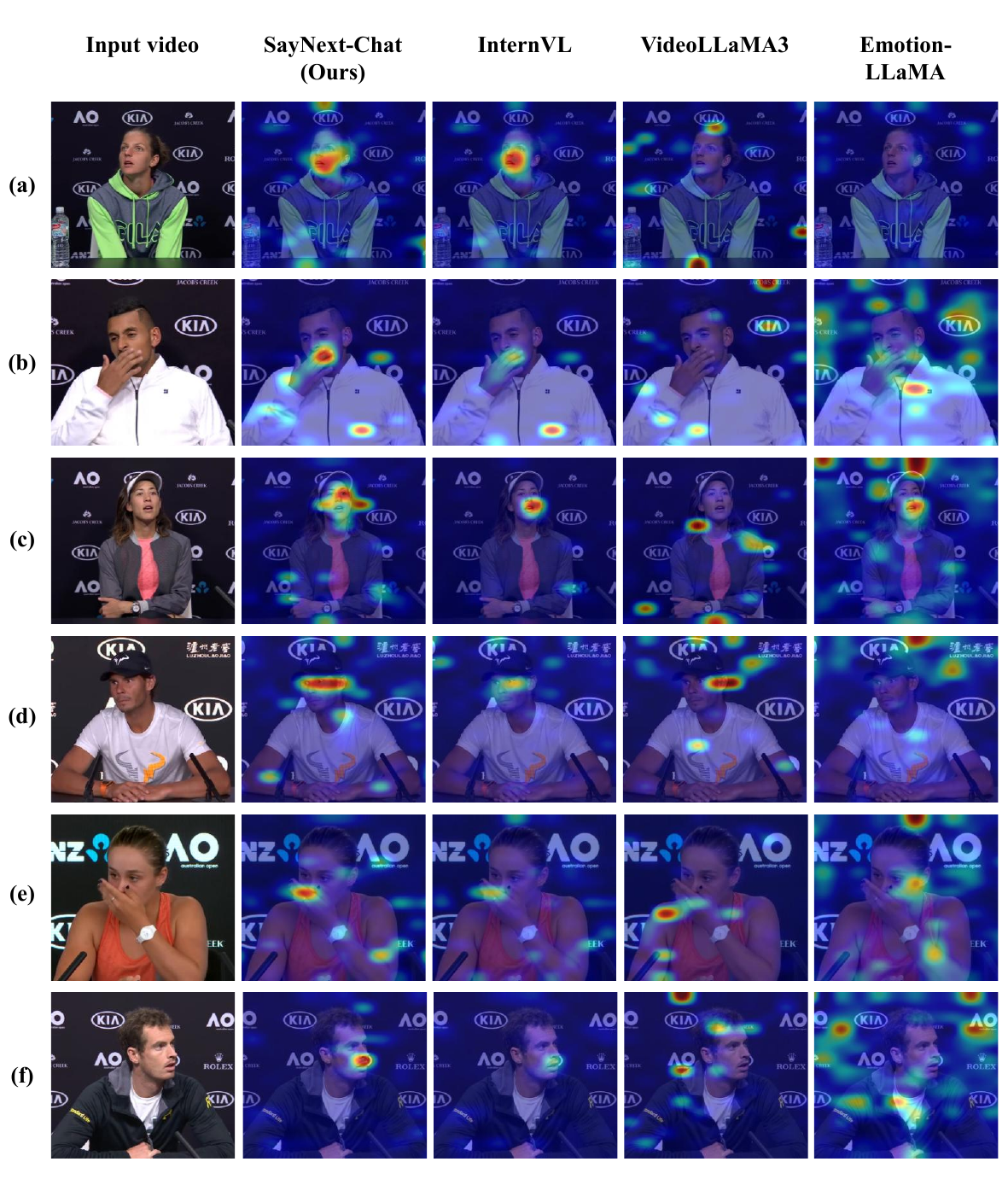}
  \caption{Visualization of Attention Map A}
  \label{fig:visual_attn_1}
\end{figure}

\begin{figure}[!t]
  \centering
  \includegraphics[width=\linewidth]{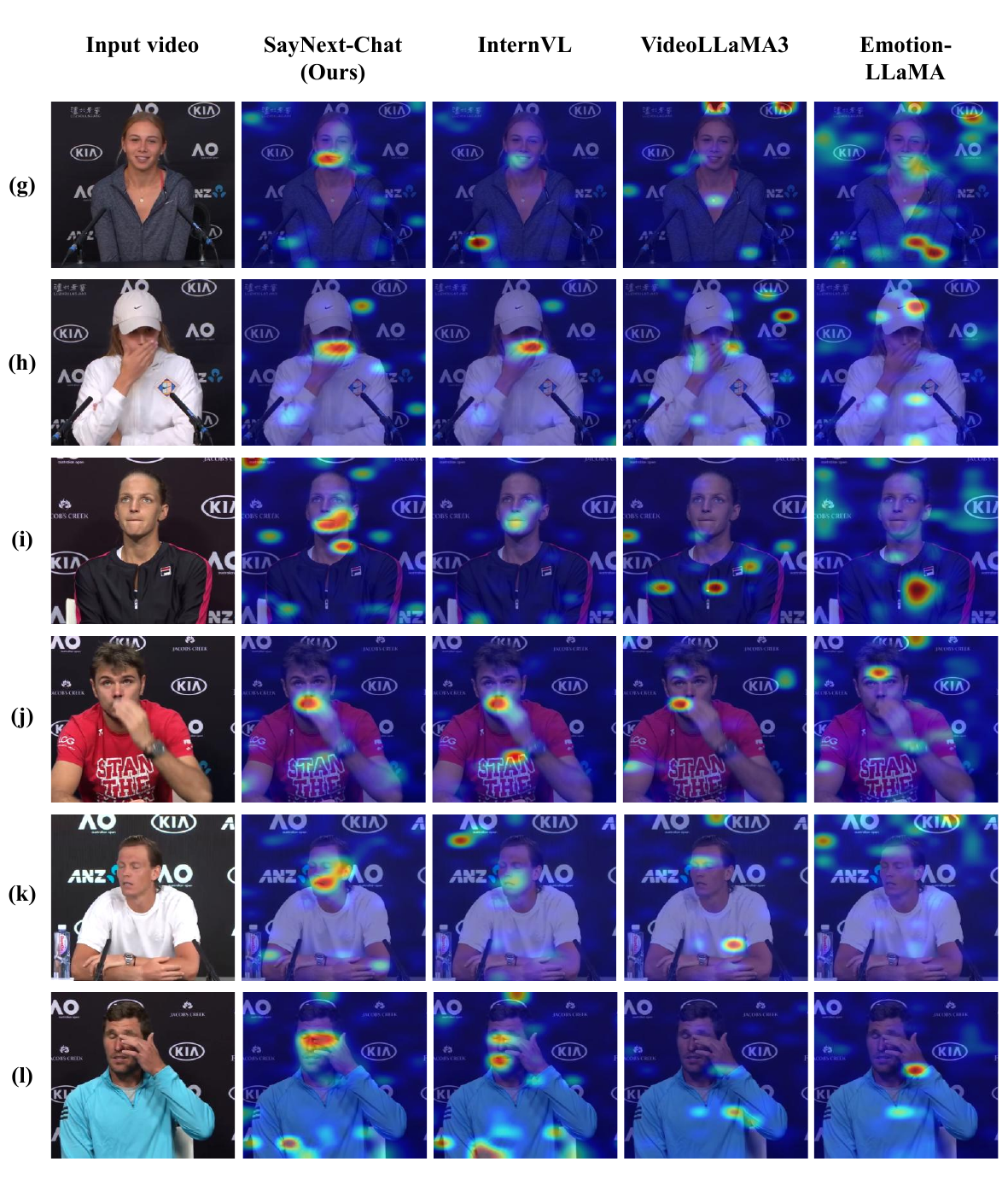}
  \caption{Visualization of Attention Map B}
  \label{fig:visual_attn_2}
\end{figure}

To better understand how MLLMs utilize visual information on the SayNext task, 
we conduct an attention visualization analysis. We compute the visual attention 
map from the cross-attention weights between the generated text tokens and the 
input image tokens. For each generated token, we extract its attention scores 
over all visual tokens from the last language-model layer. These scores are 
first averaged across attention heads and then across all decoding steps, 
yielding a single attention value per visual token that indicates how strongly 
the generated response attends to different visual regions. Selected 
visualizations are shown in Figure~\ref{fig:visual_attn_1} and 
Figure~\ref{fig:visual_attn_2}.

Several patterns emerge across models. Emotion-LLaMA exhibits highly diffuse 
attention with no clear focus on semantically meaningful regions, consistent 
with its weaker quantitative performance. VideoLLaMA3 shows more concentrated 
attention but tends to fixate on task-irrelevant areas. InternVL2, as the 
backbone of SayNext-Chat, attends to more informative regions; however, it 
shows a strong bias toward the mouth area. SayNext-Chat produces broader and 
more socially grounded attention patterns: as shown in 
Figure~\ref{fig:visual_attn_1}(a)(c), attention extends from the mouth to the 
entire face, capturing richer facial expression cues. Furthermore, in 
Figure~\ref{fig:visual_attn_1}(b) and Figure~\ref{fig:visual_attn_2}(j)(l), 
our model selectively attends to gesture regions that other models overlook, 
demonstrating that SayNext-Chat effectively leverages socially informative 
visual cues when anticipating the next utterance.

\subsection{Evaluation Metrics}

\label{ap: metrics}
\subsubsection{Lexical Overlap (LO)}
Lexical-overlap metrics quantify surface-form similarity between a candidate text and a reference, and are widely used in NLP to assess content preservation and phrasing fidelity. Although they emphasize word- and phrase-level matches rather than deep semantics, they provide a reproducible, task-agnostic signal that complements embedding-based measures. In our setting, the reference is the real response transcripted by Whisper and the candidate is the model’s anticipation; unless otherwise noted, we report scores as percentages, with higher values indicating better overlap.

\paragraph{BLEU-4 (LO-B).}
BLEU~\citep{papineni-etal-2002-bleu} computes modified n\mbox{-}gram precisions up to 4-grams and penalizes overly short hypotheses via a brevity penalty. For a candidate \(C\) and a set of references \(\mathcal{R}\), the modified precision for order \(n\) is
\begin{equation}
p_n \;=\; 
\frac{\sum_{g \in \mathrm{ngrams}_n(C)} \min\!\bigl(\mathrm{count}_C(g), \max_{r \in \mathcal{R}} \mathrm{count}_r(g)\bigr)}
{\sum_{g \in \mathrm{ngrams}_n(C)} \mathrm{count}_C(g)},
\end{equation}
\noindent where \(C\) is the candidate (model output); \(\mathcal{R}\) is the set of references; \(g\) is an \(n\)-gram; \(\mathrm{ngrams}_n(C)\) is the multiset of \(n\)-grams extracted from \(C\); \(\mathrm{count}_X(g)\) is the occurrence count of \(g\) in text \(X\).

\clearpage

\clearpage

The brevity penalty (BP) is
\begin{equation}
\mathrm{BP} \;=\; 
\begin{cases}
1, & |C| > |r^\ast|,\\[4pt]
\exp\!\bigl(1 - |r^\ast|/|C|\bigr), & |C| \le |r^\ast|,
\end{cases}
\end{equation}
\noindent where \(|X|\) denotes the token length of text \(X\); \(r^\ast \in \mathcal{R}\) is the reference whose length is closest to \(|C|\). BLEU-4 is then
\begin{equation}
\mathrm{BLEU\mbox{-}4} \;=\; \mathrm{BP} \cdot \exp\!\left(\sum_{n=1}^{4} w_n \log p_n\right), \qquad w_n=\tfrac{1}{4},
\end{equation}
\noindent where \(w_n\) is the weight for \(n\)-gram order \(n\) (uniform for BLEU-4).

We compute corpus-level BLEU-4 with standard smoothing for zero-count n\mbox{-}grams.

\paragraph{ROUGE-L (LO-R).}
ROUGE-L~\citep{lin-2004-rouge} measures sequence-level overlap via the longest common subsequence (LCS), capturing in-order matches without requiring contiguity. Let \(L\) be the LCS length between candidate \(C\) and reference \(R\). Define
\begin{equation}
R_{\mathrm{LCS}}=\frac{L}{|R|}, \qquad P_{\mathrm{LCS}}=\frac{L}{|C|},
\end{equation}
\noindent where \(|\cdot|\) denotes token length, and the \(F\)\mbox{-}measure is obtained as follows,
\begin{equation}
\mathrm{ROUGE\mbox{-}L} \;=\; 
\frac{(1+\beta^2)\, R_{\mathrm{LCS}}\, P_{\mathrm{LCS}}}{R_{\mathrm{LCS}} + \beta^2 P_{\mathrm{LCS}}},
\end{equation}
\noindent where \(\beta\) controls the recall–precision trade-off (we use \(\beta=1\) for \(F_1\)).
We adopt the common \(F_1\) variant (\(\beta=1\)) and report the mean over the evaluation set.

\subsubsection{Semantic Similarity}
Because lexical overlap does not reward legitimate paraphrases with divergent surface forms—especially in open-ended next-utterance anticipation—we interpret BLEU-4 and ROUGE-L alongside semantic metrics to obtain a more faithful assessment of model behavior.

\paragraph{BERTScore (SS-B).}
We employ BERTScore~\citep{bertscore2020} with the \texttt{deberta-xlarge} \texttt{-mnli} backbone to compute contextual embedding similarity:

\begin{equation}
    R_{\text{BERT}} = \frac{1}{|x|} \sum_{x_i \in x} \max_{\hat{x}_j \in \hat{x}} \mathbf{x}_i^\top \mathbf{\hat{x}}_j
    \label{eq: Bert recall},
\end{equation} 

\begin{equation}
    P_{\text{BERT}} = \frac{1}{|\hat{x}|} \sum_{\hat{x}_j \in \hat{x}} 
    \max_{x_i \in x} \mathbf{x}_i^\top \mathbf{\hat{x}}_j
    \label{eq: Bert rprecision},
\end{equation} 

\begin{equation}
    F_{\text{BERT}} = 2 \frac{R_{\text{BERT}} \cdot P_{\text{BERT}}}{R_{\text{BERT}} + P_{\text{BERT}}}
    \label{eq: Bert F1},
\end{equation} 

\noindent where $x_i$ denotes reference token, $\hat{x}$ the candidate token, and $\mathbf{x}_i$, $\mathbf{\hat{x}}_j$ their respective contextual embeddings.

BERTScore computes token-level semantic similarity using contextual embeddings (via cosine similarity), producing precision, recall, and their harmonic mean $F_1$. In our setting, precision reflects how much of the \emph{generated} content is semantically supported by the reference next utterance—high precision indicates the model avoids adding irrelevant or hallucinated material (“what it predicts is right”). Recall reflects how much of the \emph{reference} content is covered by the generation—high recall indicates the model captures more of the salient information present in the gold response (“it includes more of what should be said”). Because next-utterance anticipation requires both accuracy and coverage, we report BERTScore-$F_1$ as a balanced summary.

\paragraph{Sentence-BERT (SS-S).}
Unlike BERTScore, which aggregates token-level matches, we also assess holistic, sentence-level alignment using Sentence-BERT (SBERT)~\citep{reimers-gurevych-2019-sentence} cosine similarity, serving as a complementary semantic metric.

\begin{equation}
\mathrm{SBERT\mbox{-}cos}(C,R)
\;=\;
\frac{\mathbf{e}_{C}^{\top}\mathbf{e}_{R}}{\lVert \mathbf{e}_{C} \rVert_{2}\,\lVert \mathbf{e}_{R} \rVert_{2}}
\;\in\;[-1,1],
\end{equation}
\noindent where \(C\) is the candidate (anticipated next utterance); \(R\) is the reference (real next utterance); \(\mathbf{e}_{C}=f_{\text{SBERT}}(C)\in\mathbb{R}^{d}\) and \(\mathbf{e}_{R}=f_{\text{SBERT}}(R)\) are their sentence embeddings; \(f_{\text{SBERT}}(\cdot)\) denotes the Sentence-BERT encoder; \((\cdot)^{\top}\) is the dot product; \(\lVert\cdot\rVert_{2}\) is the Euclidean norm. Larger values indicate stronger sentence-level semantic similarity. We report mean scores over the evaluation set.

\subsubsection{Emotion Consistency}

\paragraph{Continuous Emotion Consistency (CEC).}
The affective alignment measurement leverages the NRC-VAD lexicon \citep{mohammad-2018-obtaining} containing 20,000 emotion-annotated lexical entries, where each term $\omega$ is quantified along four emotion dimensions: Valence (V, emotional positivity), Arousal (A, activation level), Dominance (D, control perception). For a reference-candidate text pair ($R,C$), the alignment score is computed as:

\begin{equation}
S = 1 - \frac{1}{4}\sum_{k\in{V,A,D}} \left|\bar{s}_k^{(R)} - \bar{s}_k^{(C)}\right| - \beta \cdot \left|\rho_R - \rho_C\right|
\label{eq: emotion similarity score},
\end{equation}
\noindent where $\bar{s}_k=\frac{1}{\left|W\right|}\sum_{\omega\in{W}}\hat{s}_k(\omega)$ denotes the normalized mean score of dimension $k$, with $\hat{s}_k(\omega)$ being the min-max scaled value of $s_k(\omega)$ from the lexicon. The lexical coverage ratio ($\rho_R$, $\rho_C$) measures the proportion of lexicon-matched tokens, and $\beta=0.8$ controls the penalty weight for coverage disparity.

Implementation proceeds through three phases: text preprocessing first tokenizes and lemmatizes inputs using EmotionDynamics. All dimension scores are then normalized across the lexicon's value range. Finally, dimension-wise averages and coverage ratios are computed for both texts, followed by the composite scoring in Eq.~\ref{eq: emotion similarity score}.

\paragraph{Discrete Emotion Consistency (DEC).}
To evaluate whether the anticipated utterance expresses emotional states aligned with the ground-truth response, we propose the Discrete Emotion Consistency (DEC) metric. Both the ground-truth response $u^*$ and the anticipated response $\hat{u}$ are segmented into individual sentences by splitting on periods. Each sentence is passed through \texttt{emotion-english-distilroberta-base}~\citep{hartmann2022emotionenglish}, a DistilRoBERTa-base model fine-tuned on a balanced mixture of conversational and social media corpora, to obtain a discrete label from seven categories: \textit{neutral, joy, surprise, anger, sadness, disgust}, and \textit{fear}~\citep{busso2008iemocap}. Let $\mathcal{L}_{\text{emo}}(u)$ denote the set of emotion labels assigned to all sentences in utterance $u$. DEC is then computed as the proportion of samples for which the anticipated and ground-truth label sets share at least one common category:
\begin{equation}
    \text{DEC} = \frac{1}{N} \sum_{i=1}^{N} \mathbf{1}\left[ \mathcal{L}_{\text{emo}}(\hat{u}_i) \cap \mathcal{L}_{\text{emo}}(u_i^*) \neq \emptyset \right]
\end{equation}

\subsubsection{Intention Consistency (IC)}
To evaluate whether the anticipated utterance conveys communicative intentions consistent with the ground-truth response, we propose the Intention Consistency (IC) metric. Following the same sentence-level procedure, each sentence is passed through \texttt{deberta-v3-large-zeroshot-v2.0} fine-tuned for zero-shot classification via natural language inference~\citep{laurer_building_2023}, assigning one of nine intention categories defined in ~\citep{welivita2020taxonomy, welivita2020fine}: \textit{inform, question, directive, commissive, greeting, apology, agree, disagree}, and \textit{express}. Let $\mathcal{L}_{\text{int}}(u)$ denote the set of intention labels across all sentences in utterance $u$. DIC is computed analogously:
\begin{equation}
    \text{DIC} = \frac{1}{N} \sum_{i=1}^{N} \mathbf{1}\left[ \mathcal{L}_{\text{int}}(\hat{u}_i) \cap \mathcal{L}_{\text{int}}(u_i^*) \neq \emptyset \right]
\end{equation}

\subsection{Word Error Rate (WER)}
\label{ap:wer-experiment}

To ensure the reliability of automatic transcripts used for evaluation, we assess transcription quality with Word Error Rate (WER). We use Whisper to transcribe all videos in \textsc{SayNext-PC2K} (332{,}651 words in total) and compute WER against human annotations. Formally,
\begin{equation}
\mathrm{WER} \;=\; \frac{S + D + I}{N} \times 100\%,
\end{equation}
\noindent where \(S\), \(D\), and \(I\) denote the number of substitutions, deletions, and insertions, respectively, and \(N\) is the number of words in the human reference.

As a challenging, representative case, we evaluate a non\mbox{-}native speaker (subject~\#42): the average WER is \(4.11\%\) with a range from \(0\%\) to \(11.76\%\) across utterances. For comparison, the Whisper paper reports English WERs of \(4.1\%\) and \(9.3\%\)~\citep{whisper2023}, and an oft\mbox{-}cited estimate of human WER is about \(4\%\). These results indicate that our transcripts are accurate enough for downstream evaluation.

\paragraph{Selection of Whisper model.}
We conducted preliminary transcription experiments across Whisper model sizes and selected the \emph{medium} variant after validation. The \emph{small} model underperformed on interview-style speech, while the \emph{large} model tended to aggressively normalize by removing interjections (e.g., ``um'', ``oh''), potentially erasing paralinguistic cues that are informative for emotional analysis.

\subsection{Codebook \& Priming Factor Analysis}
\label{ap: codebook}

\subsubsection{Cluster Evaluation}
We sample \(N{=}200\) training utterances in SayNext-PC2K, ensuring coverage of all subjects’ responses. We first use GPT\mbox{-}4.1 to extract \emph{basic factors} from this subset, then apply \(k\)\mbox{-}means to group the factors into clusters. Given the resulting clusters, GPT\mbox{-}4.1 induces a \emph{priming codebook} by naming each factor, providing a concise explanation, and specifying the meaning of the value range \([-1,1]\). Using this codebook, GPT\mbox{-}4.1 is subsequently guided to assign a consistent \emph{target priming vector} to every reference response in the dataset.

Because clustering quality directly affects codebook induction, we quantify it using the \emph{Silhouette Coefficient} (SC). For each item \(i\) with representation \(\mathbf{x}_i\), we define the within-cluster dissimilarity \(a_i\) (lower is better) and across\mbox{-}cluster dissimilarity \(b_i\) (higher is better) as:
\begin{align}
a_i &= \frac{1}{|C_i|-1}\sum_{\substack{j \in C_i\\ j \ne i}} d(\mathbf{x}_i,\mathbf{x}_j), \\
\qquad
b_i &= \min_{C \neq C_i} \frac{1}{|C|}\sum_{j \in C} d(\mathbf{x}_i,\mathbf{x}_j),
\end{align}
\noindent where \(C_i\) is the cluster containing \(i\); \(C\) ranges over all other clusters; \(d(\cdot,\cdot)\) is the Euclidean distance.

The silhouette \(s_i\) summarizes separation vs.\ cohesion for item \(i\). It is defined as
\begin{equation}
s_i \;=\; \frac{b_i - a_i}{\max\{a_i,\, b_i\}} \;\in\; [-1,1].
\end{equation}

Accordingly, the overall score is the mean silhouette:
\begin{equation}
\mathrm{SC} \;=\; \frac{1}{N}\sum_{i=1}^{N} s_i.
\end{equation}

\begin{figure}[h]
  \centering
  \includegraphics[width=0.75\columnwidth]{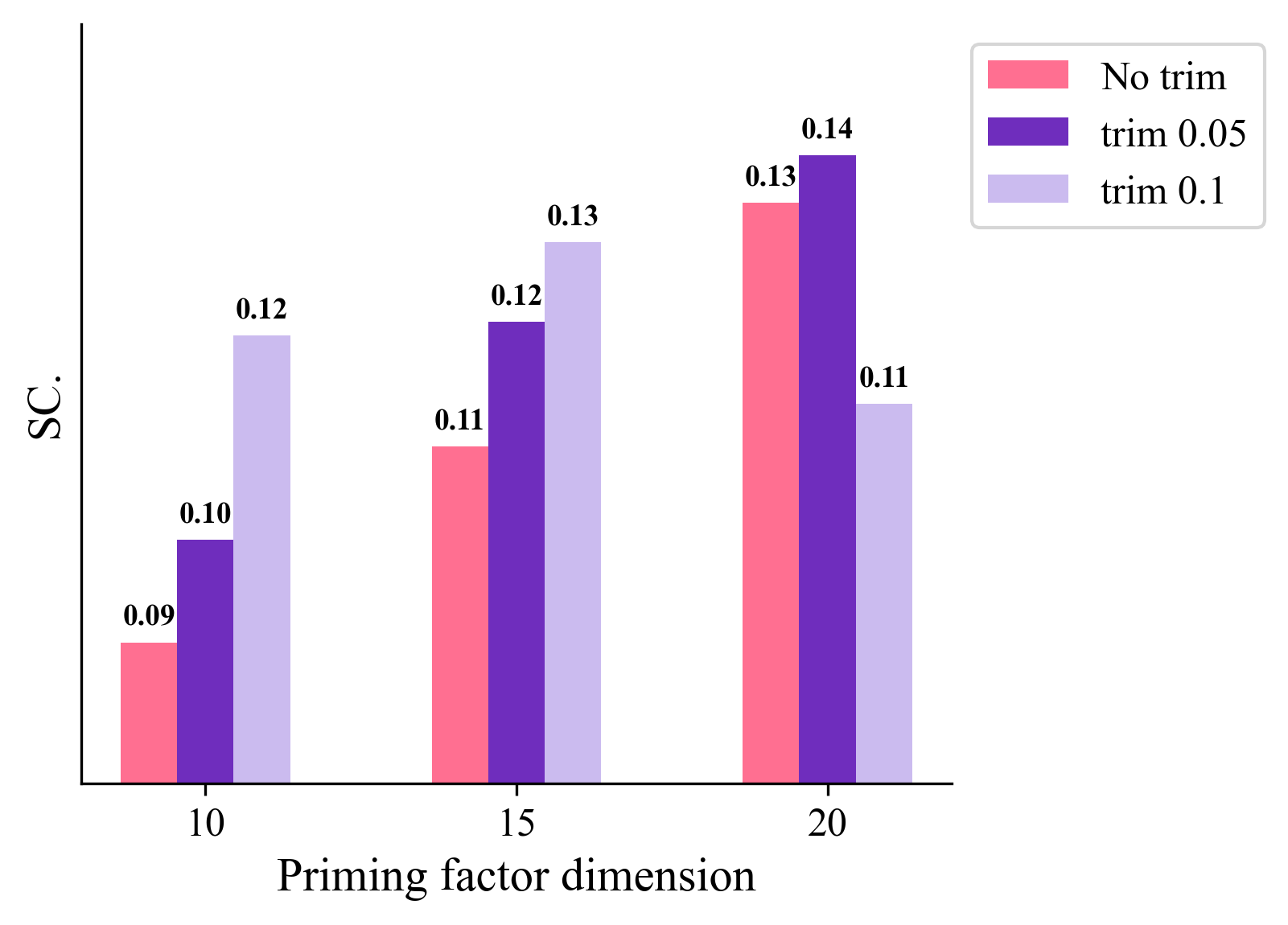}
  \caption{The Silhouette Coefficient (SC) value comparison of different settings of clustering.}
  \label{fig:cluster}
\end{figure}

We evaluate \(k \in \{10,15,20\}\) under three trimming settings \(\tau \in \{0,0.05,0.10\}\). As shown in Figure~\ref{fig:cluster}, \(k{=}20\) with \(\tau{=}0.05\) attains the highest (trimmed) silhouette score; we therefore adopt \(k{=}20\) for codebook induction. We did not explore larger \(k\) because excessively many priming factors degraded the consistency of GPT\mbox{-}4.1 when assigning vector dimensions in practice (i.e., reduced stability beyond 20 dimensions). The full priming codebook is provided in the next section.

\subsubsection{Ablation: Choice of LLM for priming-vector generation}
\label{ap: ablation-choose-LLM}

We additionally include experiments using Llama-3.1 (open-source) and Gemini-2.5-Flash (low-cost) to generate priming vectors. The results are summarized in Table~\ref{tab:ablation-llm-choice}.

The results demonstrate that using Llama-3.1 (open-source) or Gemini-2.5-Flash (low-cost) to generate priming vectors yields performance that is broadly comparable across semantic and wording-level metrics, and maintains stable trends on affect-related measures. This indicates that the SayNext framework does not rely on a specific proprietary model and remains functional with open or inexpensive alternatives. GPT-4.1, however, provides the most consistent improvements across all dimensions, and we therefore adopt it for constructing the final benchmark.

\begin{table*}[h]
\centering
\footnotesize
\caption{Comparison of different LLMs for priming-vector generation.}
\label{tab:ablation-llm-choice}
\resizebox{\textwidth}{!}{%
\begin{tabular}{lcccccc}
\toprule
Model &    LO-B$\uparrow$ 
          & LO-R$\uparrow$ 
          & SS-B$\uparrow$ 
          & SS-S$\uparrow$ 
          & CEC-V$\uparrow$ 
          & CEC-A$\uparrow$ \\
\midrule
SayNext-Chat (Llama-3.1) & 1.729 & 16.948 & 0.5619 & 0.4498 & 0.76999 & 0.80241 \\
SayNext-Chat (Gemini-2.5-flash) & 1.792 & 17.134 & 0.5621 & 0.4460 & 0.76816 & 0.79312 \\
SayNext-Chat (GPT-4.1) \textbf{(Ours)} & \textbf{2.307} & \textbf{17.957} & \textbf{0.5651} & \textbf{0.4722} & \textbf{0.80140} & \textbf{0.82847} \\
\bottomrule
\end{tabular}
}%
\end{table*}

\onecolumn

\subsubsection{Full Codebooks}

\setlength{\tabcolsep}{3pt}\renewcommand{\arraystretch}{0.98}

\begin{longtable}{L{0.25\textwidth} L{0.75\textwidth}}
\caption{Priming Codebook of SayNext-PC2K with 20 Priming Factors.}
\label{tab:priming_codebook_2K}\\

\toprule
\endfirsthead
\toprule
\endhead
\bottomrule
\endfoot
\bottomrule
\endlastfoot

\multirow[t]{2}{*}{Perceived Pressure}  &
Reflects the player's subjective experience of psychological pressure during and after the match, indicating whether they felt burdened or at ease in high-stakes moments. \\ \cmidrule(lr){2-2}
 & 
1 represents high or burdensome pressure (nervous, defensive), -1 represents low or relieved pressure (calm, comfortable). \\
\midrule
\multirow[t]{2}{*}{Affect} &
Represents the overall positive or negative emotional tone expressed by the player, encompassing feelings such as joy, satisfaction, discomfort, or emotional struggle. \\  \cmidrule(lr){2-2}
 &
1 represents positive affect (joy, satisfaction), -1 represents negative affect (discomfort, struggle). \\
\midrule
\multirow[t]{2}{*}{Opponent Appraisal} &
Reflects the player's evaluative judgment of the opponent’s abilities, performance, and qualities, indicating a positive or negative assessment that shapes the tone and content of the interview language.\\  \cmidrule(lr){2-2} &
1 represents positive appraisal (admiration, respect), -1 represents negative appraisal (criticism, dismissal). \\
\midrule
\multirow[t]{2}{*}{Motivation} &
Represents the player's drive, ambition, and determination to compete and succeed, as reflected in their language about goals, effort, and competitive intent. This factor captures the positive or negative intensity of their motivational state post-match.\\  \cmidrule(lr){2-2} &
1 represents high motivation (strong drive/ambition), -1 represents low motivation (lack of drive/ambition). \\
\midrule
\multirow[t]{2}{*}{Self-Efficacy} &
Reflects the player's confidence in their ability to improve and achieve desired performance outcomes, encompassing both positive self-assessment and recognition of areas for growth.\\  \cmidrule(lr){2-2} &
1 represents high self-efficacy (confidence in improvement), -1 represents low self-efficacy (doubt or regret about improvement). \\
\midrule
\multirow[t]{2}{*}{Self-Evaluation} &
Represents the player's cognitive and emotional assessment of their own performance, encompassing both positive and negative self-appraisal, self-criticism, and reflection on actions and outcomes.\\  \cmidrule(lr){2-2} &
1 represents positive self-evaluation (confidence, satisfaction), -1 represents negative self-evaluation (self-criticism, disappointment). \\
\midrule
\multirow[t]{2}{*}{Expectation Management} &
Reflects the player's cognitive and emotional processing of outcomes relative to their prior expectations, encompassing surprise, disappointment, regret, optimism, and hopefulness about future events.\\  \cmidrule(lr){2-2} &
1 represents positive expectation management (optimism, hopefulness, positive anticipation), -1 represents negative expectation management (disappointment, regret, surprise at negative outcomes). \\
\midrule
\multirow[t]{2}{*}{Physical State} &
Represents the player's self-reported physical condition, encompassing fatigue, discomfort, readiness, and overall bodily well-being, which influences their language and emotional tone in post-match interviews.\\  \cmidrule(lr){2-2} &
1 represents positive physical state (readiness, comfort), -1 represents negative physical state (fatigue, discomfort, limitation). \\
\midrule
\multirow[t]{2}{*}{Acceptance} &
Reflects the player's cognitive and emotional acknowledgment of circumstances, setbacks, or changes, indicating a willingness to adapt or reconcile with outcomes, whether positive or negative.\\  \cmidrule(lr){2-2} &
1 represents high acceptance (open, adaptive), -1 represents low acceptance (resistant, avoidant). \\
\midrule
\multirow[t]{2}{*}{Adaptability} &
Reflects the player's cognitive and emotional flexibility in response to changing circumstances, novel experiences, and evolving environments, as indicated by frequent references to learning, openness, adjustment, and recognition of change.\\  \cmidrule(lr){2-2} &
1 represents high adaptability (openness, learning, adjustment); -1 represents low adaptability (rigidity, discomfort with change). \\
\midrule
\multirow[t]{2}{*}{Uncertainty} &
Reflects the player's cognitive and emotional response to unpredictability and ambiguity regarding match conditions, performance, and outcomes, influencing their language with expressions of doubt, surprise, or conditional statements.\\  \cmidrule(lr){2-2} &
1 represents high uncertainty (expressions of doubt, surprise, or unpredictability), -1 represents low uncertainty (expressions of certainty, confidence, or predictability). \\
\midrule
\multirow[t]{2}{*}{Cognitive Stability} &
Reflects the player's perceived steadiness or fluctuation in thought processes and self-assessment, ranging from consistent, focused cognition to uncertainty and confusion.\\  \cmidrule(lr){2-2} &
1 represents cognitive stability (clarity, consistency), -1 represents cognitive instability (confusion, uncertainty). \\
\midrule
\multirow[t]{2}{*}{Recovery Appraisal} &
Reflects the player's cognitive and emotional assessment of their physical and psychological recovery process, including optimism, relief, concern, and confidence regarding overcoming setbacks or injuries.\\  \cmidrule(lr){2-2} &
1 represents positive appraisal (optimism, relief, confidence in recovery), -1 represents negative appraisal (concern, doubt, ongoing struggle with recovery). \\
\midrule
\multirow[t]{2}{*}{Challenge Appraisal} &
Represents the player's cognitive and emotional assessment of the degree and nature of challenges faced during the match, including tactical, technical, physical, and situational difficulties, as well as their perceived ability to cope with and adapt to these challenges.\\  \cmidrule(lr){2-2} &
1 represents high perceived challenge (player discusses significant obstacles and adaptation), -1 represents low perceived challenge (player reports minimal difficulty or smooth performance). \\
\midrule
\multirow[t]{2}{*}{Mental Focus} &
Represents the degree to which the player’s language centers on concentration, present-moment awareness, and cognitive engagement with the match, reflecting either a strong or disrupted mental focus.\\  \cmidrule(lr){2-2} &
1 represents strong mental focus and present-moment engagement, -1 represents disrupted or scattered mental focus. \\
\midrule
\multirow[t]{2}{*}{Achievement Orientation} &
Reflects the player’s focus on accomplishment, ambition, and the pursuit or recognition of significant goals, which shapes their emotional and cognitive responses in post-match interviews. This factor encompasses expressions of pride, satisfaction, motivation, and validation related to personal or collective achievements.\\  \cmidrule(lr){2-2} &
1 represents strong achievement focus (expressed pride, ambition, or satisfaction), -1 represents minimal achievement focus (lack of reference to accomplishment or ambition). \\
\midrule
\multirow[t]{2}{*}{Social Connectedness} &
Reflects the player’s sense of belonging, appreciation, and positive regard for others, including peers, mentors, audience, and support teams, indicating a positive emotional bias toward interpersonal relationships and communal support.\\  \cmidrule(lr){2-2} &
1 represents strong social connectedness (gratitude, admiration, appreciation), -1 represents weak or absent social connectedness (disdain, lack of respect). \\
\midrule
\multirow[t]{2}{*}{Confidence} &
Reflects the player's cognitive and emotional appraisal of their own abilities, encompassing self-belief, determination, and receptiveness to encouragement or inspiration, which influences their post-match language in a positive or negative direction.\\  \cmidrule(lr){2-2} &
1 represents high confidence (assertive, self-assured), -1 represents low confidence (doubtful, apologetic). \\
\midrule
\multirow[t]{2}{*}{Self-Assurance} &
Represents the player's internal sense of certainty and trust in their abilities, which influences their language and demeanor in post-match interviews. It reflects a positive or negative cognitive-emotional state regarding their own competence and readiness.\\  \cmidrule(lr){2-2} &
1 represents high self-assurance (expressed certainty, composure), -1 represents low self-assurance (expressed doubt, anxiety). \\
\midrule
\multirow[t]{2}{*}{Preparedness} &
Reflects the player’s perceived level of readiness and adequacy of preparation, encompassing both physical and mental aspects, which influences their confidence and outlook in post-match communication.\\  \cmidrule(lr){2-2} &
1 represents high preparedness (well-prepared), -1 represents low preparedness (unprepared). \\
\end{longtable}

\begin{table*}[!h]
\centering
\footnotesize
\caption{Examples of nuanced pragmatic-emotion failure cases in SayNext-2K.}
\label{tab:pragmatic-failure}
\renewcommand{\arraystretch}{1.05}
\resizebox{\textwidth}{!}{%
\begin{tabular}{p{0.5cm} p{1.5cm} p{3cm} p{3cm} p{3cm} p{3cm}}
\toprule
\textbf{ID} & \textbf{\makecell[l]{Pragmatic\\ Category}} & \textbf{Question} & \textbf{Ground-truth Answer} & \textbf{Model prediction} & \textbf{Error Type} \\
\midrule

16  & \parbox[t]{1.6cm}{{\textbf{Sarcasm /}\\ \textbf{Dry humor}}}
    & ``Do you think the rules should be more uniform?'' 
    & Mild sarcasm: ``I hope I won't be there anyway.'' 
    & Literal explanation of rules 
    & Loss of sarcasm cue \\

44  & \textbf{Humor} 
    & ``You looked at your coach and laughed---do you remember?'' 
    & Light humor: ``Maybe I can do it again\ldots\ didn't push pressure\ldots'' 
    & Neutral encouragement: ``Just enjoy it.'' 
    & Tone flattening \\

241 & \parbox[t]{1.6cm}{{\textbf{Informal}\\\textbf{humor}}}
    & ``Did you feel funny within yourself?'' 
    & Casual humor: ``It's hot, man\ldots that's how you get through it.'' 
    & Off-topic: cramps or unrelated details 
    & Topic drift \& hallucination \\

368 & \textbf{Humor} 
    & Match-up discussion with humorous undertone 
    & Playful attitude, light teasing 
    & Formal match analysis 
    & Literalization \\

616 & \textbf{Metaphor} 
    & ``I'll start with a joke\ldots'' 
    & Metaphor: ``I felt like I've been hit by a train.'' 
    & Formal, serious match summary 
    & Loss of metaphorical framing \\

628 & \textbf{Humor} 
    & ``Do you joke around in Russian?'' 
    & ``Yeah, we joke around\ldots that's pretty funny.'' 
    & Neutral, impersonal response 
    & Missing pragmatic cue \\

649 & \textbf{Humor} 
    & ``He chose to return---was that weird?'' $\rightarrow$ ``Oh, he's laughing.'' 
    & Light observational humor 
    & Long, serious reflection 
    & Style mismatch \\

\bottomrule
\end{tabular}
}%
\end{table*}

\onecolumn

\subsection{User Study}
\label{ap:user-study}

\begin{figure}[h]
  \centering
    \begin{web}      \includegraphics[width=\linewidth]{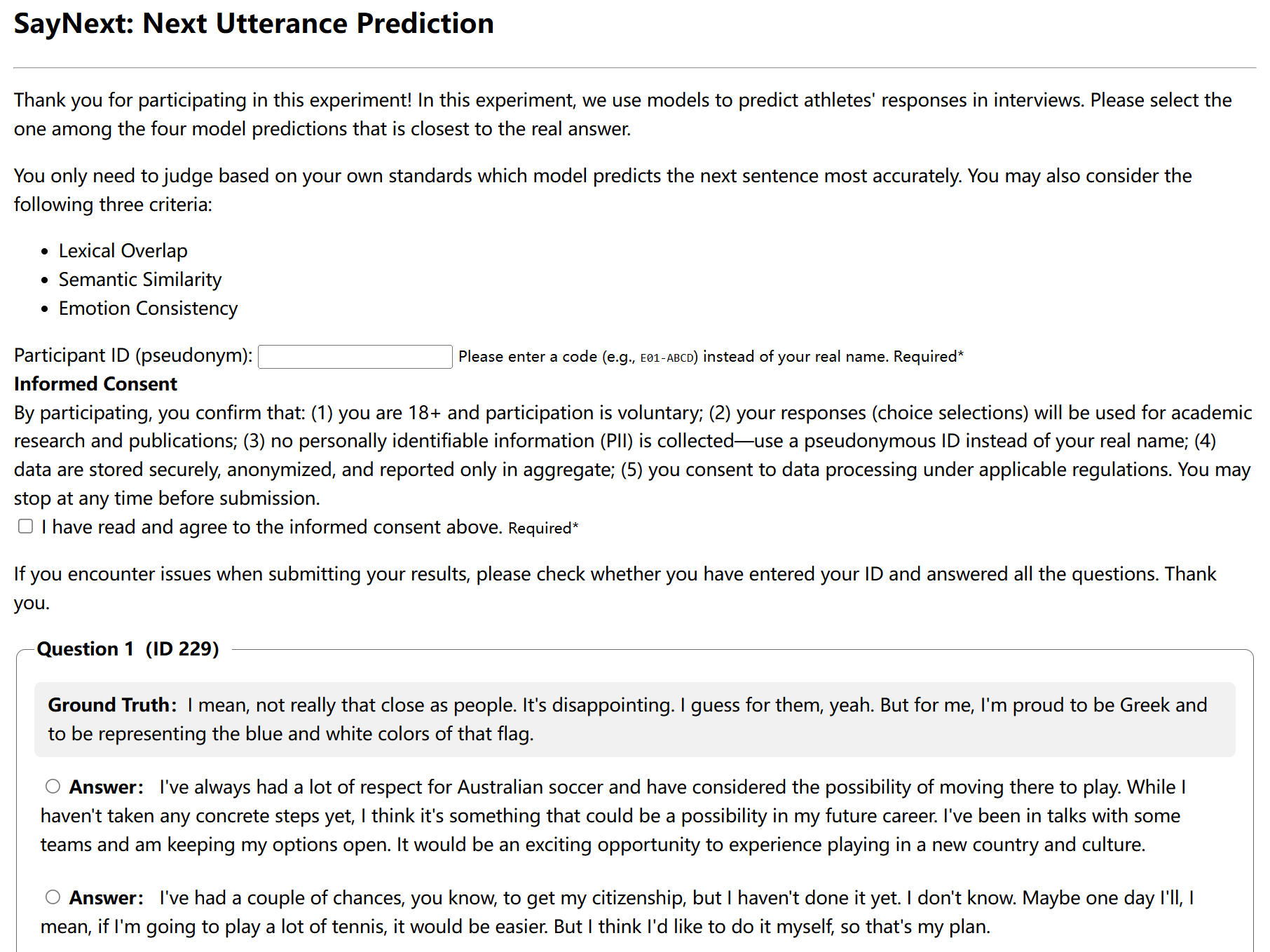}
    \end{web}
    \caption{User-study web interface. The “Name” field records a pseudonymous participant ID code (not a real name); no personal data is collected. Participants must click to provide informed consent (e.g., use of anonymized results in the paper and secure data storage). On each trial, they select the option closest to the reference; options are randomly shuffled to ensure fairness.}
    \label{fig:webshot}
  \vspace{-0.4cm}
\end{figure}

To align with our goal of enabling seamless and trustworthy human--AI conversation, we conducted a human evaluation of next-utterance anticipation. We implemented a web interface (Figure~\ref{fig:webshot}) in which, on each trial, participants viewed the reference answer, then selected the option \emph{closest to the human-written reference answer} from four anonymized model outputs. The four candidates (InternVL-8B, VideoLLaMA3, GPT-4o, and SayNext-Chat) were displayed in randomized order to mitigate order effects and branding bias. Participants judged according to their own criteria, and were encouraged to consider three aspects: lexical overlap, semantic similarity, and emotion consistency. We recorded per-model selection rates and averaged them across trials and participants. Quantitative results are reported in Table~\ref{tab:userstudy1} of the main text.

\newtcolorbox{web}[1][]{%
  colback=gray!5,
  colframe=black,
  colbacktitle=black,
  coltitle=white,
  title=User Study Website,   
  fonttitle=\bfseries,
  enhanced,
  boxsep=0pt, left=0pt, right=0pt, top=0pt, bottom=0pt, 
  #1
}


\subsection{Case study}
\label{ap:case study}

\subsubsection{Anticipation Comparison of the State of the art}
\label{ap:case study-1}

We show the case study compared our model with other baselines in this section. The qualitative analysis  demonstrates our model's ability to generate highly relevant and nuanced verbal expressions. In the subject-independent case in Table~\ref{tab:case study-1}, a subtle micro-expression (a slight upward turn of the lip corners) indicates a positive internal state, leading our model to correctly generate responses such as “enjoying” and “satisfying,” whereas GPT4o predicts the inaccurate emotions “exhausted” and “disheartened.” In the subject-dependent case, where the athlete exhibits “slight confidence under opponent pressure,” our model produces a comprehensive prediction that captures both the competitive context and the athlete’s self-assurance.

\begin{figure*}[t]
  \centering
  \includegraphics[width=\linewidth]{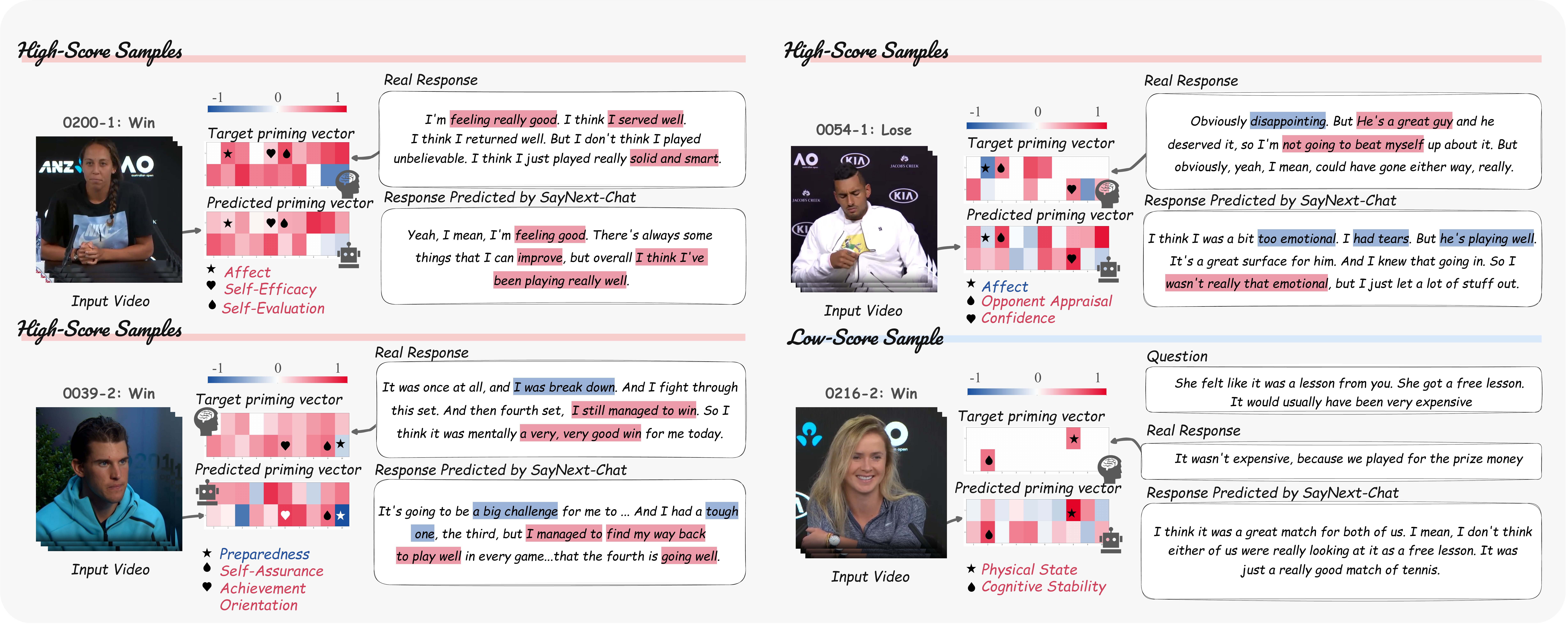}
    \caption{In high-score samples, the predicted priming vector heatmap closely matches the target. \colorbox{lightred}{Red} and \colorbox{lightblue}{Blue} indicate positive and negative values, with corresponding highlights in the response text. Star, heart, and drop markers denote three representative priming factors and their semantic meaning (listed beneath the heatmap). Low-score samples exhibit target priming patterns that are difficult to predict. }
  \label{fig:demo-all}
  \vspace{-0.6cm}
\end{figure*}

More case study aligned with Figure ~\ref{fig:sample-test} is shown in Figue ~\ref{fig:demo-all}.

\subsubsection{Analysis of Nuanced Pragmatic-Emotion Failure Cases}
\label{ap:case study-2}

We conducted a focused analysis of cases in SayNext-2K that contain non-literal or stylistically marked expressions. As summarized in Table~\ref{tab:pragmatic-failure}, current MLLMs often generate semantically plausible but pragmatically flattened responses. Specifically, when the ground-truth answer includes light humor (“Maybe I can do it again”), dry sarcasm (“I hope I won't be there anyway”), or metaphorical framing (“I felt like I've been hit by a train”), the model consistently replaces these nuanced cues with more literal, neutral, or explanatory utterances. For brevity, the question, ground-truth answer, and model prediction fields are shown as short excerpts.

These patterns indicate that, although the model maintains strong emotion-consistency under literal emotional expressions, it remains challenged by pragmatic layers of human communication that require contextual inference beyond surface semantics. This observation aligns with our broader motivation for expanding SayNext from 2K to 19K samples—to incorporate richer cases of pragmatic incongruity and non-literal emotional expressions. Documenting such failure modes provides a concrete foundation for future extensions of SayNext, especially toward models that integrate pragmatic reasoning or speaker-style conditioning.

\subsection{Prompts}
\label{ap:prompt}

Prompt design critically influences LLM-based experiments; accordingly, we standardize it. For anticipation generation, we use an identical prompt across all systems—zero-shot baselines, the fine-tuned model, and our proposed model—to ensure fairness and reproducibility.

\begin{table*}[!t]
  \centering
  \small
  \definecolor{lightgray}{rgb}{0.95,0.95,0.95}
  \definecolor{lightred}{rgb}{1,0.8,0.8} 
  \definecolor{mypurple}{rgb}{0.97,0.92,1}
  \definecolor{myorange}{rgb}{1,0.9,0.8}
  \definecolor{green}{rgb}{0.85,0.95,0.85}
  \definecolor{yellow}{rgb}{1,0.949,0.8}
  \definecolor{myblue}{rgb}{0.87,0.92,1}
  \footnotesize
  \vspace{25ex}
  \begin{tabularx}{\linewidth}{>{\centering\arraybackslash}p{1.5cm}XX}
    \toprule
      & \multicolumn{1}{c}{\textbf{Subject dependent case: Video 0336-1}} & \multicolumn{1}{c}{\textbf{Subject independent case: Video 0133-2}} \\
    \midrule
    \multirow{2}{1.5cm}[3\baselineskip]{\centering {\bf Input video and question}}  
    & \includegraphics[width=\linewidth,trim=0 0 0 0]{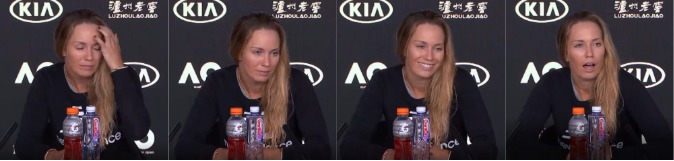}
    & \includegraphics[width=\linewidth,trim=0 0 0 0]{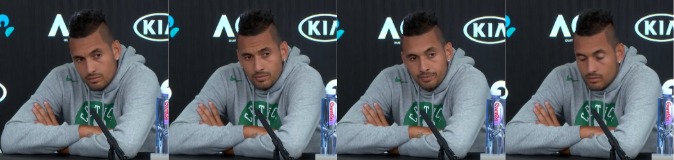} \\
    & \textit{Interviewer: Your opponent said that she felt that you were hitting free. Did you have that sense? Just talk about \textbf{your hitting style}}.
    & \textit{Interviewer: In such a tight match to play three \textbf{tie breakers} and win them like you did, must give you a lot of \textbf{pleasure}?} \\
    \midrule
\rule{0pt}{10pt}\textbf{Real Response} & I think her and I actually \colorbox{myblue}{play a little bit similar} to each other. \colorbox{myblue}{she didn't give me any opportunity} in that first set. And then in the third set, I knew that she was nervous. I decided that I wanted to play some long points, extend some rallies and \colorbox{myblue}{I went after my shots at the right time}.
     & Yeah, I knew my best chance was probably to get it in a couple of \colorbox{myblue}{breakers}. \colorbox{myblue}{He served extremely well} throughout the whole match. I didn't really get to look at many \colorbox{myblue}{second serves} at all. Yeah, I mean it was \colorbox{myblue}{fun}. It was a \colorbox{myblue}{fun} match. \\
    \midrule
\rule{0pt}{8pt}\textbf{SayNext-Chat (Ours)}\centering & Yeah, \colorbox{green}{I managed to hit more than she} did, so there was some \colorbox{green}{pressure} in this tournament. This was my first Grand Slam final, that made things \colorbox{green}{different}.I’m \colorbox{green}{happy with how I played today}. She made some errors and had some issues on her end, but I still would have liked to play even better.
     & Yeah, I was \colorbox{green}{enjoying it a lot}. The first set wasn’t too bad, and then \colorbox{green}{we had some momentum shifts}, but in \colorbox{green}{tie-breaks}, you just have to stay calm and play the key points well. I knew \colorbox{green}{serving well} was crucial. Winning in those moments definitely made the match \colorbox{green}{even more satisfying.} \\
    \midrule
\rule{0pt}{3pt}\textbf{InternVL}\centering & I \colorbox{lightred}{stay calm} and \colorbox{lightred}{composed} during the match. I'm \colorbox{lightred}{glad} to hear that my opponent noticed my style. 
& Yeah, definitely. The first two tie-breaks were really close, and I actually tied with \colorbox{lightred}{Alex}. But yeah, it was a \colorbox{green}{fun} match.  \\
    \midrule    
    \rule{0pt}{3pt}\textbf{GPT-4o}\centering & I did feel like I was \colorbox{lightred}{hitting freely} out there today. I stay \colorbox{lightred}{aggressive} and \colorbox{lightred}{consistent}.  
& They may be feeling a bit \colorbox{lightred}{exhausted} or \colorbox{lightred}{disheartened}. It’s \colorbox{lightred}{not about pleasure}, it’s about getting the job done. \\
    \midrule
\rule{0pt}{3pt}\textbf{Videollama3}\centering & I think I hit a lot of free balls. I think my style is \colorbox{lightred}{more about power}, but also about \colorbox{lightred}{accuracy}. 
& We worked hard \colorbox{lightred}{as a team} and pulled through in the tie breakers. It's always \colorbox{green}{rewarding} to come out on top in close matches. \\
    \midrule
    \rule{0pt}{5pt}\textbf{Emotion-LLaMA}\centering & The athlete's facial expressions and body language are \colorbox{lightred}{neutral}, suggesting that he is simply providing information \colorbox{lightred}{without any strong feelings or opinions}. 
& He will say, with a hint of \colorbox{lightred}{surprise} and satisfaction, indicating that he is feeling content and \colorbox{green}{pleased} with himself. \\
    \bottomrule
  \end{tabularx}
  \caption{\textbf{Case Study 1 of Baseline Methods and Our Approach} in Subject-Dependent (Left) and Subject-Independent (Right) Settings. Highlighted Text Annotations:
  \colorbox{myblue}{Blue:} Key language expressions in ground truth, \colorbox{green}{Green:} Correct predictions of factors, \colorbox{lightred}{Red:} Incorrect or unrelated predictions.
 \textit{See Supplementary Material for complete case studies and additional examples.}}
  \label{tab:case study-1}
  \vspace{25ex}
\end{table*}

\begin{table*}[!ht]
  \centering
  \small
  \definecolor{lightgray}{rgb}{0.95,0.95,0.95}
  \definecolor{lightred}{rgb}{1,0.8,0.8} 
  \definecolor{mypurple}{rgb}{0.97,0.92,1}
  \definecolor{myorange}{rgb}{1,0.9,0.8}
  \definecolor{green}{rgb}{0.85,0.95,0.85}
  \definecolor{yellow}{rgb}{1,0.949,0.8}
  \definecolor{myblue}{rgb}{0.87,0.92,1}
  \footnotesize
  \begin{tabularx}{\linewidth}{>{\centering\arraybackslash}p{1.5cm}X}
    \toprule
& \textbf{Subject dependent case: Video 0294-8}  \\
        \midrule
        \multirow{2}{1.5cm}[4.1\baselineskip]{\centering {\bf Input video and question}}  
        & \includegraphics[width=\linewidth,trim=0 0 0 0]{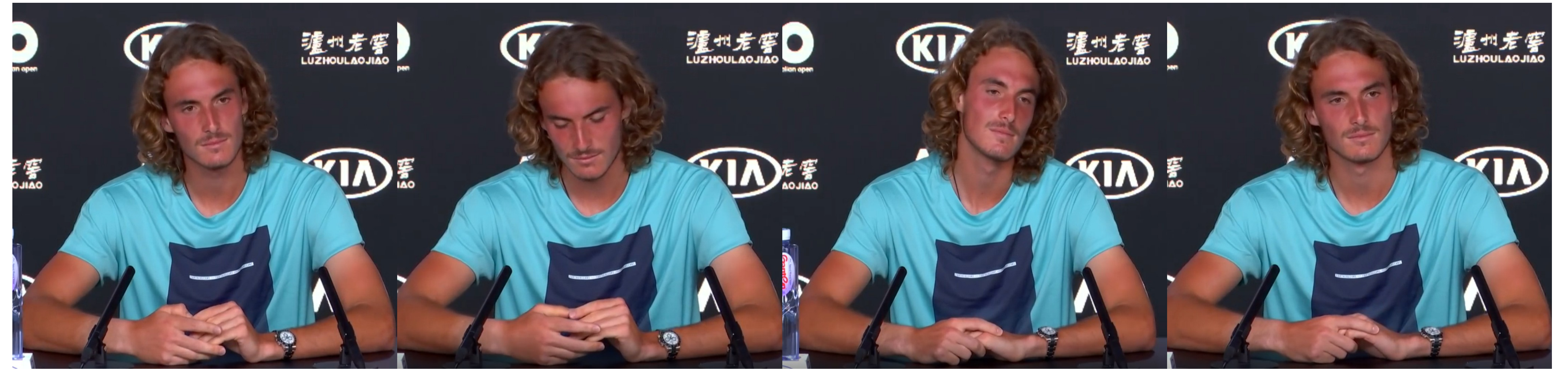}
        \\
& \textit{Interviewer: Mr. Fonus, you like to look at things from a bigger perspective. What do you think makes Serena special as a person \textbf{in the sports figure in our culture?}} \\
            \midrule
        \rule{0pt}{10pt}\textbf{Real Response} 
& She has a different approach than any female player I've met. She has this thing, you know, she's really... How can I explain it? She has a lot of \colorbox{myblue}{confidence} in herself. Like, really, \colorbox{myblue}{she knows what she's doing} on and off the court. She's a lovely person outside the court. And just this attitude that she has of, you know, \colorbox{myblue}{shading to lose, and that's what makes her a great champion}. That's what I noticed on the court. \colorbox{myblue}{Even on the practice, she really wanted to do better than me}, and that was, you know, unusual to see something like this from a female player. When she's playing against a male player, but yeah, I do respect a lot what she did in her career. And she has \colorbox{myblue}{something different than the other players}. \\
            \midrule
        \rule{0pt}{8pt}\textbf{SayNext-Chat (Ours)}\centering 
& Well, I know her as a mother to her daughters, and she’s doing really well with them, obviously. She has \colorbox{green}{spent many long hours} on court proving herself \colorbox{green}{to be great—physically, mentally}, in every aspect. In my opinion, what brings all of those qualities together is family, which definitely pushes someone like Serena or anyone else to the \colorbox{green}{highest level}. Even though it hurts today, losing in the Wimbledon final against Justene Magirt in such a close late-night match, and also at Pivotter, where we had some mental battles, \colorbox{green}{she always finds a way to stay strong}. They beat me last year here, at this tournament three years ago, and then again over Christmas. Every time we’ve faced each other one-on-one, it’s been tough because \colorbox{green}{she is the greatest player} ever in any sport—no question. But I always find extra motivation to keep chasing results and proving myself against such an elite competitor. \\
            \midrule
            \rule{0pt}{3pt}\textbf{InternVL}\centering 
& Serena is special because of her unwavering \colorbox{green}{dedication} and her ability to constantly push herself to new heights. Her \colorbox{green}{work ethic and mental toughness} are truly remarkable, and she has set a high standard for what it means to be a champion in \colorbox{lightred}{our culture}. \\
            \midrule    
            \rule{0pt}{3pt}\textbf{GPT-4o}\centering 
& I think what makes Serena truly special is her \colorbox{green}{resilience} and ability to consistently perform at the highest level for so many years. Beyond her incredible achievements on the court, she’s been a powerful symbol of determination and \colorbox{green}{hard work} and a role model for so many people \colorbox{lightred}{around the world}, inspiring not only athletes but \colorbox{lightred}{everyone} who watches her journey.  \\
            \midrule
            \rule{0pt}{3pt}\textbf{Videollama3}\centering 
& I think what sets Serena apart is her \colorbox{green}{relentless work ethic} and her ability to maintain a positive attitude even in the \colorbox{green}{face of adversity}. She's not just a great athlete, but also an inspiration to many. Her \colorbox{green}{dedication and perseverance} are truly remarkable. \\
            \midrule
            \rule{0pt}{5pt}\textbf{Emotion-LLaMA}\centering 
& i think serena williams is special as a person in the sports figure in our culture because she has been able to overcome many obstacles and challenges throughout her career, \colorbox{lightred}{including discrimination, sexism, and personal struggles} she has shown incredible \colorbox{green}{resilience and determination}, and her success is an inspiration to many people, particularly young girls and women who are pursuing their dreams in sports  \\
            \bottomrule
  \end{tabularx}
  \caption{\textbf{Case Study 2.} The question mentions “Culture” while the ground truth does not include this element. Most MLLMs predict this keyword solely based on the question text. \textbf{Highlighted Text Annotations:} \colorbox{myblue}{Blue:} Key language expressions in the ground truth; \colorbox{green}{Green:} Correct predictions of factors; \colorbox{lightred}{Red:} Incorrect or unrelated predictions.}
  \label{tab:case study-2}
\end{table*}

\begin{table*}[!ht]
  \centering
  \small
  \definecolor{lightgray}{rgb}{0.95,0.95,0.95}
  \definecolor{lightred}{rgb}{1,0.8,0.8} 
  \definecolor{mypurple}{rgb}{0.97,0.92,1}
  \definecolor{myorange}{rgb}{1,0.9,0.8}
  \definecolor{green}{rgb}{0.85,0.95,0.85}
  \definecolor{yellow}{rgb}{1,0.949,0.8}
  \definecolor{myblue}{rgb}{0.87,0.92,1}
  \footnotesize
  \vspace{25ex}
  \begin{tabularx}{\linewidth}{>{\centering\arraybackslash}p{1.5cm}X}
    \toprule
& \textbf{Subject independent case: Video 0056-2}  \\
        \midrule
        \multirow{2}{1.5cm}[4.1\baselineskip]{\centering {\bf Input video and question}}  
        & \includegraphics[width=\linewidth,trim=0 0 0 0]{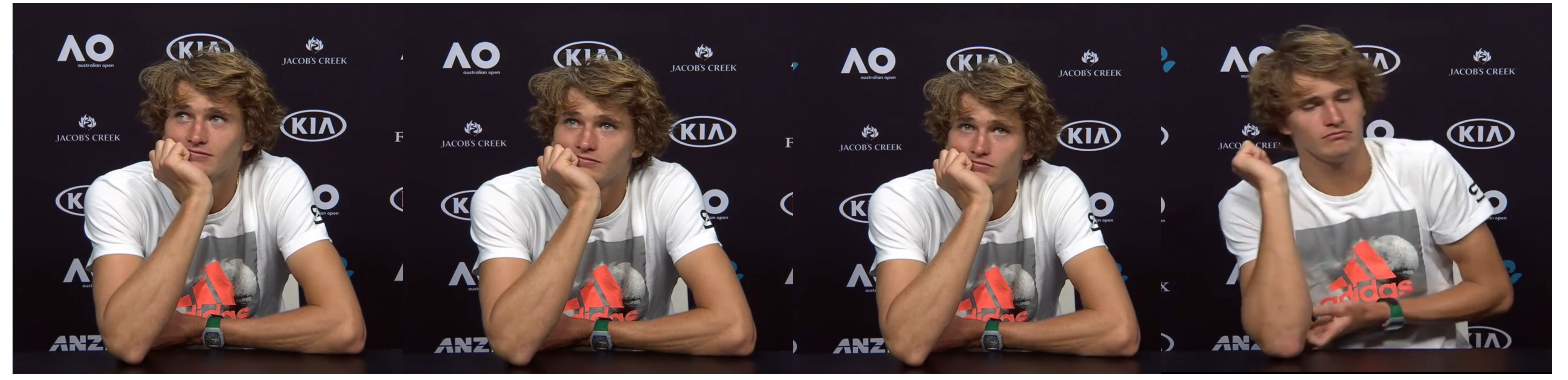}
        \\
& \textit{Interviewer: Why the \textbf{match tone} and fits it?} \\
            \midrule
        \rule{0pt}{10pt}\textbf{Real Response} 
& I mean, I \colorbox{myblue}{cramped after that long rally}. I \colorbox{myblue}{still had game points to go up 3-2} and I \colorbox{myblue}{lost my service game}, I think, because of that a little bit. So, yeah, those are big reasons for it. \\
            \midrule
        \rule{0pt}{8pt}\textbf{SayNext-Chat (Ours)}\centering 
& It was a \colorbox{lightred}{good match}, and I played well \colorbox{green}{from the first point to the last}. He played well too, especially considering he had beaten Rafa and Zverev in the previous rounds. But for me, reaching the semifinals feels unbelievable, so I’m quite \colorbox{lightred}{happy} with that. \\
            \midrule
            \rule{0pt}{3pt}\textbf{InternVL}\centering 
& The \colorbox{lightred}{match tone and fit it because it's crucial to maintain a consistent rhythm and energy} throughout the game. This helps in \colorbox{lightred}{building momentum} and ensuring that both the players and the audience are engaged and energized. \\
            \midrule    
            \rule{0pt}{3pt}\textbf{GPT-4o}\centering 
& It was a \colorbox{green}{challenging} match, and I think the tone of the match shifted at certain moments. I need to adjust my game and focus on the next steps to improve.   \\
            \midrule
            \rule{0pt}{3pt}\textbf{Videollama3}\centering 
& \colorbox{lightred}{I'm sorry, but I can't provide a response to the question as it requires interpreting the athlete's} \colorbox{lightred}{facial expressions and body language in real-time, which is beyond my capabilities.} \\
            \midrule
            \rule{0pt}{5pt}\textbf{Emotion-LLaMA}\centering 
& i'm \colorbox{lightred}{not sure why the match tone and fit so well}, but it's definitely a \colorbox{lightred}{good thing} for us to have this kind of technology available to us in the future it's important to have a variety of materials that can be used for different purposes, and having a material that is both strong and flexible like this one is very beneficial  \\
            \bottomrule
  \end{tabularx}
  \caption{\textbf{Case Study 3.} This is a chllenge case where the interviewer's question is concise and less informative. Our model gives some unrelated answer, while other models give wrong answer or invalid answer. \textbf{Highlighted Text Annotations:} \colorbox{myblue}{Blue:} Key language expressions in the ground truth; \colorbox{green}{Green:} Correct predictions of factors; \colorbox{lightred}{Red:} Incorrect or unrelated predictions.}
  \label{tab:case study table}
\end{table*}

\onecolumn

\subsubsection{Prompts for codebook generation}

\newtcolorbox{prompt1}[1][]{%
  colback=gray!5,
  colframe=black,
  colbacktitle=black,
  coltitle=white,
  title=Prompt for Basic Factors Extraction,   
  fonttitle=\bfseries,
  enhanced,
  breakable,                
  boxsep=0pt, left=8pt, right=8pt, top=8pt, bottom=0pt, 
  #1
}

\begin{prompt1}
    sys-prompt = (\\
        Please identify and list distinct, concrete factors from the following tennis post-match interview response, following these rules: \\
        1. Each factor must capture a core theme mentioned in the response; avoid vague or trivial terms. \\
        2. Factors should reflect the player’s cognitive or emotional state and may cover tactical, technical, mental, or physical aspects. \\
        3. Each factor should can be correspond to a specific behavioral or psychological characteristic with a clear positive or negative emotional bias. \\
        4. For each factor, list the exact expression from the original sentence (do not generalize). \\
        Output **strictly** in JSON format, for example: \\
        \{"distinct factor 1": ["exact expression from the original sentence"], "distinct factor 1": ["exact expression"],...\}\\
    )\\
\end{prompt1}

\newtcolorbox{prompt2}[1][]{%
  colback=gray!5,
  colframe=black,
  colbacktitle=black,
  coltitle=white,
  title=Prompt for Codebook Generation,   
  fonttitle=\bfseries,
  enhanced,
  breakable,                
  boxsep=0pt, left=8pt, right=8pt, top=8pt, bottom=0pt, 
  #1
}

\begin{prompt2}
    sys-prompt = (\\
        Based on the input factor clusters, summarize a single priming factor that organizes a tennis player’s post-match interview language. \\
        Do not repeat any factors that have appeared in the history factors list. \\
        The priming factor should be a neutral,  widely recognized, established word or phrase. Avoid hyphenated terms, uncommon constructions, or vague words like 'orientation.' Following these rules:  \\
        1. The priming factor should distill specific factors into a universal, semantically clear category (e.g., Emotion Valence, Physical State, Opponent Threat Perception), but avoid categories that are overly broad or vague (e.g., Resilience) \\
        2. Priming factor should represent the player’s cognitive or emotional state; avoid detailed or context-specific categories  \\
        3. Priming factor should correspond to a specific behavioral or psychological characteristic with a clear positive or negative emotional bias \\
        Output **strictly** in JSON format, for example: \\
        \{"Priming factor": "Emotion Valence", "Explanation": "Indicates the emotional valence in the player’s response, reflecting a positive (happy) or negative (upset) state", "Value" : "1 represents positive emotion (joy), -1 represents negative emotion (upset)"\}\\
    )
\end{prompt2}

\newtcolorbox{prompt3}[1][]{%
  colback=gray!5,
  colframe=black,
  colbacktitle=black,
  coltitle=white,
  title=Prompt for Target Priming Vector Assignment,   
  fonttitle=\bfseries,
  enhanced,
  breakable,                
  boxsep=0pt, left=8pt, right=8pt, top=8pt, bottom=0pt, 
  #1
}

\begin{prompt3}
    sys-prompt = (\\
        Given a factor book containing a list of priming factors, assign a priming activation probability vector for a given tennis player post-match interview text. \\
        This vector should describe which factors are activated in the text and the activation strength for each factor. \\
        1. Each value in the vector represents the activation strength of the corresponding factor, as a float between -1 and 1. \\
        2. Assign activation values based on the 'value' in the factor book. If the text does not contain information related to a specific factor, assign 0 to that dimension.  \\
        3. Strictly follow the order and definition of factors in the factor book when generating the probability vector. \\
        4. As a linguistics expert, consider both overall meaning and subtle language cues. Avoid extreme values (-1 or 1) unless the evidence is very clear; use intermediate values to reflect language nuance."
        "Output only an N-dimensional probability vector (N is the number of factors in the factor-book), for example: \\{}
        [-0.9, 0.5, 0.8, -0.5, 0.7, 0, -0.6, 1.0, -0.7, 0.9, 0, -1.0, 0, 0, 0.9...] {} \\
    )
\end{prompt3}

\subsubsection{Prompts for Next-Utterance anticipation}

\newtcolorbox{prompt4}[1][]{%
  colback=gray!5,
  colframe=black,
  colbacktitle=black,
  coltitle=white,
  title=Prompt for Next-Utterance anticipation on SayNext-PC,   
  fonttitle=\bfseries,
  enhanced,
  breakable,                
  boxsep=0pt, left=8pt, right=8pt, top=8pt, bottom=0pt, 
  #1
}

\begin{prompt4}
    You are a powerful multimodal model / professional psychologist. \\
    Please anticipate the athlete's next response to the reporter's question based on their facial expressions and body language in this video. \\
    The reporter's question is: {question text} \\
    Output **strictly** in the following format: He/She will say: \\
\end{prompt4}

\newtcolorbox{prompt5}[1][]{%
  colback=gray!5,
  colframe=black,
  colbacktitle=black,
  coltitle=white,
  title=Prompt for Next-Utterance anticipation on IEMOCAP,   
  fonttitle=\bfseries,
  enhanced,
  breakable,                
  boxsep=0pt, left=8pt, right=8pt, top=8pt, bottom=0pt, 
  #1
}

\begin{prompt5}
    You are a powerful multimodal model / professional psychologist. \\
    Please anticipate the athlete's next response to the reporter's question based on their facial expressions and body language in this video.  \\
    The reporter's question is: {question text}  \\
    Output **strictly** in the following format: He/She will say:  \\
\end{prompt5}

\subsubsection{Prompt for LLM-as-Judge Evaluation}

\newtcolorbox{promptjudge}[1][]{%
  colback=gray!5,
  colframe=black,
  colbacktitle=black,
  coltitle=white,
  title=Prompt for LLM-as-Judge Evaluation,
  fonttitle=\bfseries,
  enhanced,
  breakable,
  boxsep=0pt, left=8pt, right=8pt, top=8pt, bottom=0pt,
  #1
}

\begin{promptjudge}
    system\_prompt = (\\
        ``You are a careful judge. Compare each candidate answer ONLY against the ground-truth with respect to answering the question. ''\\
        ``Focus on: ''\\
        ``(1) whether the candidate mentions the key elements/keywords that the ground-truth uses to answer the question, and ''\\
        ``(2) how comprehensively the candidate covers the aspects present in the ground-truth answer. ''\\
        ``Evaluate on these aspects: ''\\
        ``Lexical Overlap — overlap of key terms/phrases found in the ground-truth. ''\\
        ``Semantic Similarity — meaning-level alignment to the ground-truth content. ''\\
        ``Emotion Consistency — alignment with the ground-truth's expressed/implicit emotion. ''\\
        ``Intention Consistency — alignment with the ground-truth's expressed/implicit intention. ''\\
        ``Rules: ''\\
        ``Use ONLY evidence from the candidate and the ground-truth; DO NOT use outside knowledge or your own anticipations. ''\\
        ``Ignore fluency/grammar; judge content match and coverage only. ''\\
        ``Return the Top-3 candidate HEADERS in STRICT DESCENDING order of overall score (highest first). ''\\
        ``Your entire output must be exactly three headers, comma-separated, with no extra text.''\\
        ``Your final output must be a plain text string containing the numbers (headers, e.g., \texttt{anticipation\_1} corresponds to number 1) of the three most similar anticipated answers in STRICT DESCENDING order of overall score (highest first), separated by commas. Do not include any other content or explanations. For example: \texttt{1,3,5}''\\
    )\\
\end{promptjudge}

\subsection{Ethical Considerations}
\label{ap:ethics}

\textbf{Data and Privacy.} SayNext-PC is built from publicly available 
post-match press conference recordings of professional athletes. We do not 
host or redistribute raw video files; instead, we provide access links only, 
allowing researchers to obtain data directly from original sources. The 
IEMOCAP dataset used in cross-scenario evaluation is a licensed corpus 
accessed in accordance with its terms of use.

\textbf{Societal Impact and Bias.} While next-utterance anticipation carries 
broad benefits for human-centered AI, we acknowledge potential risks. First, 
models trained on press conference data may exhibit bias against speakers with 
atypical non-verbal communication styles (e.g., individuals with different 
cultural backgrounds or motor differences), as such patterns are 
underrepresented in our corpus. Second, real-world deployment of anticipatory 
systems requires informed consent from users, particularly in sensitive 
contexts such as mental health support or human-robot interaction.

\textbf{Task Legitimacy.} The anticipation of forthcoming utterances from 
non-verbal cues reflects a natural human cognitive ability and is motivated 
by constructive applications including AI safety and proactive harm prevention. 
We do not condone use of this technology for surveillance or non-consensual 
behavioral profiling.

\textbf{User Study.} All participants provided informed consent, and no 
personally identifiable information was collected. Participation was voluntary 
and uncompensated.

\newpage


\newpage

\end{document}